\documentclass{article} 
\usepackage{iclr2026_conference,times}

\usepackage{amsmath,amsfonts,bm}

\def\eqref#1{equation~\ref{#1}}

\def\1{\bm{1}}

\DeclareMathAlphabet{\mathsfit}{\encodingdefault}{\sfdefault}{m}{sl}
\SetMathAlphabet{\mathsfit}{bold}{\encodingdefault}{\sfdefault}{bx}{n}

\DeclareMathOperator*{\argmax}{arg\,max}

\usepackage{url}

\usepackage{amsmath}        
\usepackage{amssymb}        
\usepackage{amsthm}         
\usepackage{mathtools}      

\usepackage{algorithm}      
\usepackage{algpseudocode}  

\usepackage{booktabs}       
\usepackage{array}          
\usepackage{multirow}       
\usepackage{rotating}
\usepackage{makecell}
\usepackage[table]{xcolor}
\usepackage{minitoc}
\usepackage{titletoc}

\usepackage{longtable}
\usepackage{adjustbox} 
\usepackage{tabulary}   
\usepackage{tabularx}

\usepackage{pifont}         
\usepackage{tikz}           
\definecolor{darkgreen}{RGB}{0,128,0}
\newcommand{\cmark}{\textcolor{darkgreen}{\ding{51}}}
\newcommand{\xmark}{\textcolor{red}{\ding{55}}}

\usepackage[utf8]{inputenc}  
\usepackage[T1]{fontenc}     
\usepackage{microtype}        
\usepackage{enumitem}         
\usepackage[most]{tcolorbox}
\definecolor{pycomment}{RGB}{106,153,85}
\definecolor{pyblue}{RGB}{50,150,250}
\definecolor{pyred}{RGB}{250,80,80}
\definecolor{pygreen}{RGB}{100,200,100}
\tcbuselibrary{listingsutf8}

\usepackage{hyperref}         
\usepackage{cleveref}         

\usepackage{graphicx}         
\usepackage{xcolor}           
\usepackage{listings}         
\usepackage{subcaption}       
\usepackage{float}            
\usepackage{soul}
\usepackage{xspace}
\usepackage{wrapfig}

\theoremstyle{plain}

\theoremstyle{definition}

\newcommand{\our}{\textsc{Hindsight}\xspace}

\newcommand{\hindsight}{\textsc{Hindsight}\xspace}
\newcommand{\tempr}{\textsc{Tempr}\xspace}
\newcommand{\cara}{\textsc{Cara}\xspace}

\title{Hindsight is 20/20: Building Agent Memory that Retains, Recalls, and Reflects} 

\author{
\vspace{1em}
\hspace{-0.5em}
\textbf{Chris Latimer}$^{\clubsuit}$,
\textbf{Nicoló Boschi}$^{\clubsuit}$,
\textbf{Andrew Neeser}$^{\diamondsuit}$,
\textbf{Chris Bartholomew}$^{\clubsuit}$,\\[-1em]
\textbf{Gaurav Srivastava}$^{\heartsuit}$,
\textbf{Xuan Wang}$^{\heartsuit}$,
\textbf{Naren Ramakrishnan}$^{\heartsuit}$
\\[0.75em]
$^{\clubsuit}$Vectorize.io, USA \\
$^{\diamondsuit}$The Washington Post, USA \\
$^{\heartsuit}$Virginia Tech, USA
\\[0.75em]
}

\iclrfinalcopy 

\begin{document}

\maketitle

\begin{abstract}
Agent memory has been touted as a dimension of growth for LLM-based applications, enabling agents that can accumulate experience, adapt across sessions, and move beyond single-shot question answering.
The current generation of agent memory systems treats memory as an external layer that extracts salient snippets from conversations, stores them in vector or graph-based stores, and retrieves top-$k$ items into the prompt of an otherwise stateless model.
While these systems improve personalization and context carry-over, they still blur the line between evidence and inference, struggle to organize information over long horizons, and offer limited support for agents that must explain their reasoning.
We present \hindsight, a memory architecture that treats agent memory as a structured, first-class substrate for reasoning by organizing it into four logical networks that distinguish world facts, agent experiences, synthesized entity summaries, and evolving beliefs.
This framework supports three core operations---retain, recall, and reflect---that govern how information is added, accessed, and updated.
Under this abstraction, a temporal, entity-aware memory layer incrementally turns conversational streams into a structured, queryable memory bank, while a reflection layer reasons over this bank to produce answers and to update information in a traceable way.
On key long-horizon conversational memory benchmarks like LongMemEval and LoCoMo, Hindsight with an open-source 20B model lifts overall accuracy from 39\% to 83.6\% over a full-context baseline with the same backbone and outperforms full-context GPT-4o. Scaling the backbone further pushes Hindsight to 91.4\% on LongMemEval and up to 89.61\% on LoCoMo (vs. 75.78\% for the strongest prior open system), consistently outperforming existing memory architectures on multi-session and open-domain questions.
\end{abstract}

\doparttoc 
\faketableofcontents 

\section{Introduction}
AI agents are increasingly expected to behave less like stateless question answering systems and more like long-term partners: they are expected to remember past interactions, build up and track knowledge about the world, and maintain stable perspectives over time~\cite{packer2023memgpt,rasmussen2025zep}. However, 
the current generation of agent
memory systems 
today are still built around short-context retrieval-augmented generation (RAG) pipelines and generic large language models (LLMs). Such designs
treat memory as an external layer that extracts salient snippets
from conversations, stores them in vector or graph-based stores, and retrieves top-k
items into the prompt of an otherwise stateless model~\cite{wu2024longmemeval,maharana2024evaluating}. 

As a result, current approaches
to modeling agent memory
struggle with three recurring challenges.
First, they are  unable to preserve
and granularly access
long-term information across sessions~\cite{tavakoli2025beyond,ai2025memorybench}. Second, AI agents are
unable to epistemically
distinguish what the agent has observed from what it believes. Finally,
such agents are notorious for
their inability to exhibit preference consistency, i.e., expressing a stable reasoning style and viewpoint across interactions rather than producing locally plausible but globally inconsistent responses~\cite{huang2025llms}.

Recent work has begun to address these challenges through dedicated memory architectures for agents, e.g., see ~\cite{zhang2025survey,wu2025human}. Systems like MemGPT~\cite{packer2023memgpt} introduce operating system-like memory management, while Zep~\cite{rasmussen2025zep} proposes temporal knowledge graphs as an internal data structure. Other approaches focus on continual learning~\cite{ai2025memorybench}, reinforcement-based memory management~\cite{yan2025memory}, or production-ready memory systems~\cite{chhikara2025mem0}. 
While these systems improve personalization and context carry-over, they still blur the line between evidence and inference, can struggle to selectively organize information over long horizons, and offer limited support for agents that must explain why they answered a question a certain way.

We present \hindsight, a memory architecture for long-lived AI agents that addresses these challenges by unifying long-term factual recall with preference-conditioned reasoning. Each agent in \our is backed by a structured memory bank that accumulates everything the agent has seen, done, and decided over time, and a reasoning layer that uses this memory to answer questions, execute workflows, form opinions, and update beliefs in a consistent way. Conceptually, \our ties together two components: \tempr (Temporal Entity Memory Priming Retrieval), which
implements the \emph{retain} and \emph{recall} operations over long-term memory, and \cara (Coherent Adaptive Reasoning Agents), which implements the \emph{reflect} operation over that memory.
\tempr builds a temporal, entity-aware memory graph and exposes an agent-optimized retrieval interface, while \cara integrates configurable disposition behavioral parameters into the reasoning process and maintains an explicit opinion network that evolves over time.

At the core of \our is a simple abstraction: a memory bank organized into four logical networks (\emph{world}, \emph{experience}, \emph{opinion}, \emph{observation}) and three core operations (retain, recall, and reflect, as mentioned earlier).
The world and experience networks store objective facts about the external world and the agent's own experiences. The opinion network stores subjective beliefs with confidence scores that can be updated as new evidence arrives. The observation network stores preference-neutral summaries of entities synthesized from underlying facts. \tempr implements retain and recall by extracting narrative facts with temporal ranges, resolving entities and constructing graph links, and retrieving memories via multi-strategy search. \cara implements reflect by combining retrieved memories with an agent profile (name, background, and disposition behavioral parameters) to generate preference-shaped responses and to form and reinforce opinions.
As we will demonstrate empirically, this design provides several performance advantages over existing memory systems~\cite{xu2025mem,liu2025memverse,wang2025karma}. 


Our contributions are:
\begin{enumerate}[itemsep=-0.1em]

    \item \textbf{A unified memory architecture for agents.} \our's organization of memory into separate networks with core operations helps separate evidence, synthesize summaries better, and supports evolving beliefs, while supporting epistemic clarity and traceability.

    \item \textbf{Retain, recall and reflect layers specialized for agent memory.} Our key operational primitives help turn conversational transcripts into a structured, queryable memory bank with ability to reason over this bank and update beliefs in a stable, auditable manner.
    
    \item \textbf{Empirical evaluation on long-horizon conversational benchmarks.} We evaluate \our on LongMemEval and LoCoMo: with an open-source 20B backbone it lifts overall accuracy from 39.0\% to 83.6\% over a full-context baseline on LongMemEval and from 75.78\% to 85.67\% on LoCoMo, and with larger backbones reaches 91.4\% and 89.61\% respectively, matching or surpassing prior memory systems and frontier-backed full-context baselines.

\end{enumerate}

\section{Related Work}
\label{sec:related-work}

Recent work in agent memory spans context management systems that handle LLM window constraints, structured memory architectures built on temporal knowledge graphs, evaluation benchmarks that test these systems, and cognitive frameworks inspired by human memory. We group related work into two categories and discuss how \our differs.

\subsection{Memory Architectures and Systems}

\textbf{Tiered context management systems.} Early systems extended context using tiered architectures. MemGPT \citep{packer2023memgpt} pages information between active prompt and archival storage, treating memory as unstructured text blocks without separating facts from beliefs. LIGHT \citep{tavakoli2025beyond} handles conversations up to 10 million tokens using episodic memory, working memory, and scratchpad buffers, but does not distinguish subjective beliefs from objective observations.

\textbf{Structured memory with knowledge graphs.} Several systems use knowledge graphs for retrieval. Zep \citep{rasmussen2025zep} builds temporal knowledge graphs with bi-temporal modeling that tracks when facts are valid versus when they were recorded, but focuses on objective facts without modeling subjective beliefs or behavioral profiles. A-Mem \citep{xu2025mem} uses the Zettelkasten method to create atomic notes with LLM-generated links that evolve over time, but treats all memory uniformly without separating facts from opinions. Mem0 \citep{chhikara2025mem0} focuses on production efficiency with dense retrieval and graph representations, handling fact conflicts through database updates rather than belief evolution. Memory-R1 \citep{yan2025memory} uses reinforcement learning to train agents on memory operations to maximize QA accuracy, but does not focus on cognitive structure and behavioral profile consistency. MemVerse \citep{liu2025memverse} handles multimodal memory through a dual-path architecture combining retrieval and parametric memory via fine-tuning, which raises editability issues that \our avoids by using only external memory. KARMA \citep{wang2025karma} targets embodied AI with 3D scene graphs for spatial reasoning in robotics, not conversational agents. Table~\ref{tab:memory_systems_comparison} compares these systems across key architectural features.

\begin{table*}[ht]
\centering
\small
\begin{adjustbox}{width=0.95\textwidth,center}
\begin{tabular}{lccccccccc}
\toprule
\textbf{Feature} & \textbf{MemGPT} & \textbf{LIGHT} & \textbf{Zep} & \textbf{A-Mem} & \textbf{Mem0} & \textbf{Memory-R1} & \textbf{MemVerse} & \textbf{KARMA} & \cellcolor{gray!15} \textbf{\makecell{Hindsight\\(Ours)}}\\
\midrule
Separates facts/opinions     & \xmark & \xmark & \xmark & \xmark & \xmark & \xmark & \xmark & \xmark & \cellcolor{gray!15} \textbf{\cmark} \\
Temporal reasoning  & \xmark & \xmark & \cmark & \xmark & \xmark & \xmark & \xmark & \xmark & \cellcolor{gray!15} \textbf{\cmark} \\
Entity-aware graph & \xmark & \xmark & \cmark & \cmark & \cmark & \xmark & \xmark & \cmark & \cellcolor{gray!15} \textbf{\cmark} \\
Opinion evolution    & \xmark & \xmark & \xmark & \xmark & \xmark & \xmark & \xmark & \xmark & \cellcolor{gray!15} \textbf{\cmark} \\
Behavioral parameters     & \xmark & \xmark & \xmark & \xmark & \xmark & \xmark & \xmark & \xmark & \cellcolor{gray!15} \textbf{\cmark} \\
Confidence scores      & \xmark & \xmark & \xmark & \xmark & \xmark & \xmark & \xmark & \xmark & \cellcolor{gray!15} \textbf{\cmark} \\
External-only memory     & \cmark & \cmark & \cmark & \cmark & \cmark & \cmark & \xmark & \cmark & \cellcolor{gray!15} \textbf{\cmark} \\
Multi-strategy retrieval        & \xmark & \xmark & \xmark & \xmark & Partial & \xmark & \cmark & \xmark & \cellcolor{gray!15} \textbf{\cmark} \\
\bottomrule
\end{tabular}
\end{adjustbox}
\caption{\textbf{Comparison of memory architectures. \cmark: feature present; \xmark: absent.} \our separates objective facts from subjective opinions, maintains profile-conditioned reasoning with disposition behavioral parameters, and supports dynamic opinion evolution with confidence scores.}
\label{tab:memory_systems_comparison}
\end{table*}

\subsection{Benchmarks and Cognitive Foundations}

\textbf{Evaluation benchmarks.} Recent benchmarks that aim to evaluate long-context reasoning and selective recall have grown
in prominence.
LoCoMo~\citep{maharana2024evaluating} features very long dialogues (up to 35 sessions) wherein LLMs and traditional RAG systems struggle with long-range temporal and causal reasoning. LongMemEval~\citep{wu2024longmemeval} tests information extraction, multi-session reasoning, and temporal reasoning across conversations featuring upto 1.5 million tokens. MemoryBench \citep{ai2025memorybench} tests continual learning from feedback and finds that existing systems fail to use feedback effectively without forgetting. These benchmarks show that current systems are unable to maintain consistent behavioral profiles or handle opinion evolution.

\textbf{Cognitive foundations.} Several surveys have attempted to connect agent memory to human memory models. Recent work~\citep{zhang2025survey} categorizes memory by source, form, and operations, noting that most work focuses on task completion over consistency and that parametric memory is hard to interpret. Other work \citep{wu2025human} draws parallels between episodic/semantic memory and RAG/knowledge graphs, pointing out gaps in implicit memory and forgetting. The memory quadruple framework \citep{zhang2025memory} highlights key facets of memory---storage, persistence, access, and controllability---arguing for external memory that supports dynamic updates. Work on cognitive memory~\citep{shan2025cognitive} distinguishes explicit and implicit memory and notes that LLMs struggle with human-like knowledge integration. Recent findings \citep{huang2025llms} show that LLMs lack true working memory and must externalize state into context windows.

While earlier work has focused on storage, retrieval, and scale, recent developments in agent systems blur what agents observe versus what they believe, cannot maintain stable behavioral profiles across long interactions, and have no way to evolve subjective beliefs over time. In the rest of the paper,
we demonstrate how \our addresses these
gaps.

\section{\our Overview}
\label{sec:hindsight-overview}

\our is a memory architecture for AI agents that unifies long-term factual recall with preference-conditioned reasoning. Each agent is backed by a memory bank that accumulates interactions encountered over time, and a reasoning layer that uses this memory to answer questions, form opinions, and update its beliefs in a consistent way. 

\subsection{Four-Network Memory Organization}

At the core of \our is a memory bank organized into four logical networks, each serving a distinct epistemic role. Let $\mathcal{M} = \{\mathcal{W}, \mathcal{B}, \mathcal{O}, \mathcal{S}\}$ denote the four networks that partition the memory space, where each network maintains a specialized subset of facts. 

The \textbf{\textit{world network}} $\mathcal{W}$ stores objective facts about the external world---factual statements independent of the agent's perspective or preferences, where each fact $f_w \in \mathcal{W}$ captures information such as relationships, attributes, or events observed in the environment. 

The \textbf{\textit{experience network}} $\mathcal{B}$ stores biographical information about the agent itself, written in the first person, where each fact $f_b \in \mathcal{B}$ represents the agent's own experiences, actions, or recommendations. 

The \textbf{\textit{opinion network}} $\mathcal{O}$ stores subjective judgments formed by the agent, where each opinion $f_o \in \mathcal{O}$ is a tuple $(t, c, \tau)$ with $t$ as the opinion text, $c \in [0,1]$ as a confidence score representing belief strength, and $\tau$ as the timestamp of formation. 

The \textbf{\textit{observation network}} $\mathcal{S}$ stores preference-neutral summaries of entities synthesized from multiple underlying facts, where each observation $f_s \in \mathcal{S}$ provides a compact, objective profile derived from facts in $\mathcal{W}$ and $\mathcal{B}$.

Together, these networks provide a structured mental model of the agent's world knowledge, personal history, subjective beliefs, and synthesized entity profiles. 

\begin{figure}[t]
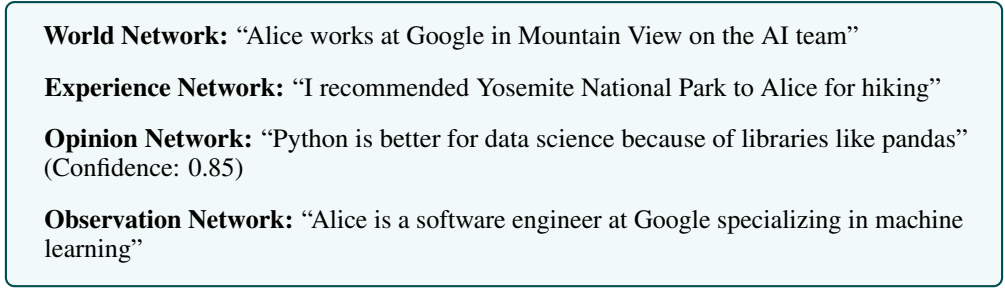

\centering
\begin{tcolorbox}[
    colback=teal!5!white,
    colframe=teal!60!black,
    fonttitle=\bfseries,
    coltitle=white,
    colbacktitle=teal!60!black,
    boxrule=0.8pt,
    arc=1mm,
    enhanced,
    width=0.95\linewidth
]
\textbf{World Network:} ``Alice works at Google in Mountain View on the AI team''

\vspace{0.3cm}

\textbf{Experience Network:} ``I recommended Yosemite National Park to Alice for hiking''

\vspace{0.3cm}

\textbf{Opinion Network:} ``Python is better for data science because of libraries like pandas'' (Confidence: 0.85)

\vspace{0.3cm}

\textbf{Observation Network:} ``Alice is a software engineer at Google specializing in machine learning''
\end{tcolorbox}

\caption{Examples of facts stored in each of the four memory networks. Each network serves a distinct epistemic role in organizing agent knowledge.}
\label{fig:four-networks}
\end{figure}

\subsection{Three Core Operations}

\our exposes the four-network memory structure through three core operations that govern how information is added, accessed, and updated. Let $B$ denote a memory bank, which is a named container that holds the four networks $\mathcal{M} = \{\mathcal{W}, \mathcal{B}, \mathcal{O}, \mathcal{S}\}$ and an associated agent profile. Let $D$ denote input data (e.g., conversational transcripts or documents to be retained), let $Q$ denote a query, and let $k$ denote a token budget. The three operations are defined as follows: 

$\text{Retain}(B, D) \rightarrow \mathcal{M}'$ takes a memory bank $B$ and input data $D$, ingests the conversational transcripts or other inputs in $D$, and converts them into narrative facts with temporal ranges, canonical entities, and graph links, extracting facts from $D$, classifying each fact into one of the four networks, and updating the memory graph (when new evidence arrives, existing beliefs in $\mathcal{O}$ are also updated through an opinion reinforcement mechanism).

$\text{Recall}(B, Q, k) \rightarrow \{f_1, \ldots, f_n\}$ takes as input memory bank $B$, query $Q$, and token budget $k$, and retrieves a variable-sized set of relevant memories from $B$ in response to query $Q$, combining semantic vector search, keyword search, graph traversal, and temporal filtering into a unified multi-strategy retrieval pipeline that returns the $n$ most relevant facts such that their combined token count does not exceed $k$.

Finally, $\text{Reflect}(B, Q, \Theta) \rightarrow (r, \mathcal{O}')$ takes memory bank $B$, query $Q$, and behavioral profile $\Theta$ (consisting of disposition behavioral parameters (skepticism, literalism, empathy) and a bias-strength parameter), generates a response $r$ to query $Q$ whose reasoning and tone are shaped by $\Theta$, first invoking recall to retrieve relevant memories from $B$, then applying preference-conditioned generation to produce a response, where new opinions may be formed during this process, resulting in an updated opinion network $\mathcal{O}'$.

\begin{figure*}[t]
    \centering
    \includegraphics[width=\textwidth]{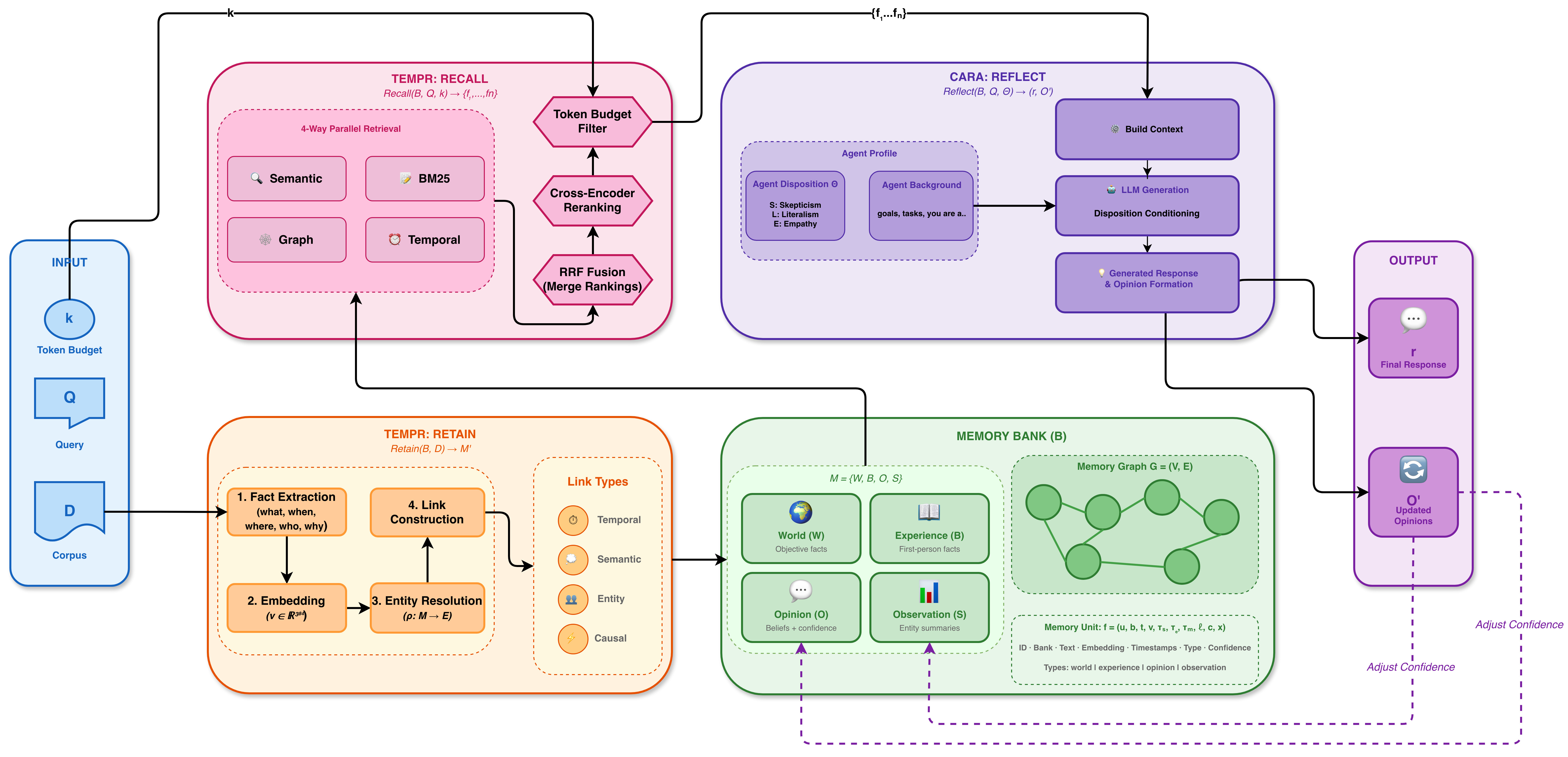}
    \caption{\textbf{End-to-end Hindsight architecture.} The system processes input data $D$ through TEMPR's retain pipeline (fact extraction, embedding generation, entity resolution, link construction) to build a structured memory bank $B$ containing four networks: world ($\mathcal{W}$), experience ($\mathcal{B}$), opinion ($\mathcal{O}$), and observation ($\mathcal{S}$). Given a query $Q$ and token budget $k$, TEMPR's recall pipeline performs four-way parallel retrieval (semantic, BM25, graph, temporal), applies Reciprocal Rank Fusion and cross-encoder reranking, and returns relevant facts. CARA's reflect operation takes these facts along with the behavioral profile $\Theta$ to generate preference-conditioned responses $r$ while forming and reinforcing opinions, updating the opinion network $\mathcal{O}'$.}
    \label{fig:hindsight-architecture}
\end{figure*}

\subsection{Component Architecture}

The two main components of \our implement these operations with distinct responsibilities:

\tempr realizes the retain and recall stages. It builds the four-network memory graph via LLM-powered narrative fact extraction, entity resolution, and link construction. TEMPR provides a retrieval interface optimized for agents, with token budgets and multi-hop discovery over temporal and entity-aware links. The retain pipeline processes input data by extracting narrative facts, generating embeddings, resolving entities, and constructing four types of graph links: temporal, semantic, entity, and causal.

\cara realizes the reflect stage. It integrates configurable disposition behavioral parameters into the reasoning process, operates over \our's networks to separate facts from beliefs, and maintains a dynamic opinion network via opinion formation and reinforcement. The behavioral profile consists of three disposition parameters (skepticism, literalism, empathy), each ranging from 1 to 5, and a bias-strength parameter between 0 and 1. CARA uses this profile to modulate the generation process, ensuring that responses align with the configured behavioral style.  Figure~\ref{fig:hindsight-architecture} provides a comprehensive view of the end-to-end architecture, showing the data flow from input through TEMPR's retain and recall pipelines, the four-network memory bank structure, and CARA's preference-conditioned reflect operation.

\subsection{Design Principles}

The architecture of \our is designed around several goals that recur throughout the paper. First, we aim for epistemic clarity, wherein
facts, observations, and opinions are kept structurally distinct so that developers and users can see what the agent knows versus what it believes. The four-network
organization
$\mathcal{M} = \{\mathcal{W}, \mathcal{B}, \mathcal{O}, \mathcal{S}\}$ provides explicit separation between objective evidence ($\mathcal{W}, \mathcal{B}$), subjective beliefs ($\mathcal{O}$), and synthesized summaries ($\mathcal{S}$). 
Second,
each memory unit $f$ carries temporal metadata $(\tau_s, \tau_e, \tau_m)$ where $\tau_s$ and $\tau_e$ define the occurrence interval and $\tau_m$ denotes the mention time, enabling precise historical queries and recency-aware ranking. This achieves
temporal awareness, wherein for a query with constraint $[\tau_{\text{start}}, \tau_{\text{end}}]$, the system retrieves facts where the occurrence interval overlaps with the query range. 
Third, this approach supports {\textit{Entity-aware reasoning}} leveraging graph links over shared entities, semantic similarity, temporal proximity, and causal relationships support multi-hop discovery of indirectly related information. The memory graph is the underlying data structure that connects all memory units: formally, $\mathcal{G} = (V, E)$ where $V$ is the set of all memory units (facts stored in the four networks) and $E$ is the set of weighted edges between them. Each edge $e \in E$ has a type $\ell \in \{\text{temporal}, \text{semantic}, \text{entity}, \text{causal}\}$ and weight $w_e \in [0,1]$, enabling traversal-based retrieval. Finally, \our aims for 
preference consistency, disposition behavioral parameters (skepticism, literalism, empathy) and a bias-strength parameter ensure that agents express stable perspectives over time while still allowing their beliefs to evolve as new evidence arrives, where the confidence score $c$ in each opinion $(t, c, \tau) \in \mathcal{O}$ is updated through a reinforcement mechanism when supporting or contradicting evidence is retained.

The following sections instantiate \our's architecture. Section~\ref{sec:tempr} describes TEMPR, which implements \our's retain and recall operations and builds the four-network memory graph. Section~\ref{sec:cara} then presents CARA, which implements the reflect operation and shows how preference-aware reasoning is layered on top of this memory substrate. Section~\ref{sec:hindsight-architecture} describes the unified integration of these components followed by experimental results.

\section{TEMPR: Retain and Recall}
\label{sec:tempr}

As described earlier, TEMPR (Temporal Entity Memory Priming Retrieval) implements \our's retain and recall operations. It is responsible for turning raw conversational transcripts into a structured, temporal, entity-aware memory graph, and for retrieving variable amounts of relevant information for downstream reasoning. 
%
We first describe how TEMPR \emph{retains} information by organizing memories, extracting narrative facts, and constructing an entity-aware graph. We then describe how it \emph{recalls} information using a four-way parallel retrieval architecture with fusion and neural re-ranking. The neural components used in this pipeline, including the embedding model for semantic retrieval, the cross encoder reranker, and the downstream LLM, can all be treated as configurable modules rather than fixed backbones.

\subsection{Retain: Building a Temporal Entity Memory Graph}
\label{sec:tempr-retain}

\subsubsection{Memory Organization}
\label{sec:memory-organization}

As introduced in Section~\ref{sec:hindsight-overview}, \our organizes memories into four networks to separate objective information, subjective beliefs, and synthesized summaries. TEMPR instantiates this design by storing each extracted fact in exactly one network and attaching it to the shared memory graph. Each fact $f$ is assigned a type $\ell(f) \in \{\text{world}, \text{experience}, \text{opinion}, \text{observation}\}$ that determines its target network.




Each memory is stored as a self-contained node that combines natural language, vector representations, and temporal metadata. Formally, a memory unit is a tuple:
\begin{equation}
f = (u, b, t, v, \tau_s, \tau_e, \tau_m, \ell, c, x)
\end{equation}
where $u$ is a unique identifier, $b$ is the bank identifier, $t$ is the narrative text, $v \in \mathbb{R}^d$ is the embedding vector, $\tau_s$ and $\tau_e$ define the occurrence interval, $\tau_m$ is the mention timestamp, $\ell$ is the fact type, $c \in [0,1]$ is an optional confidence score (for opinions), and $x$ contains auxiliary metadata such as context, access count, and full-text search vectors.

These fields allow TEMPR to treat each memory as a single unit for storage, graph construction, and retrieval, while supporting both semantic and lexical search as well as temporal and opinion-aware reasoning.

\subsubsection{LLM-Based Narrative Fact Extraction}

TEMPR uses an open-source LLM to convert conversational transcripts into narrative facts and associated metadata. Compared to rule-based or sentence-level pipelines, this approach lets us extract self-contained facts that preserve cross-turn context and reasoning.

\begin{figure}[t]
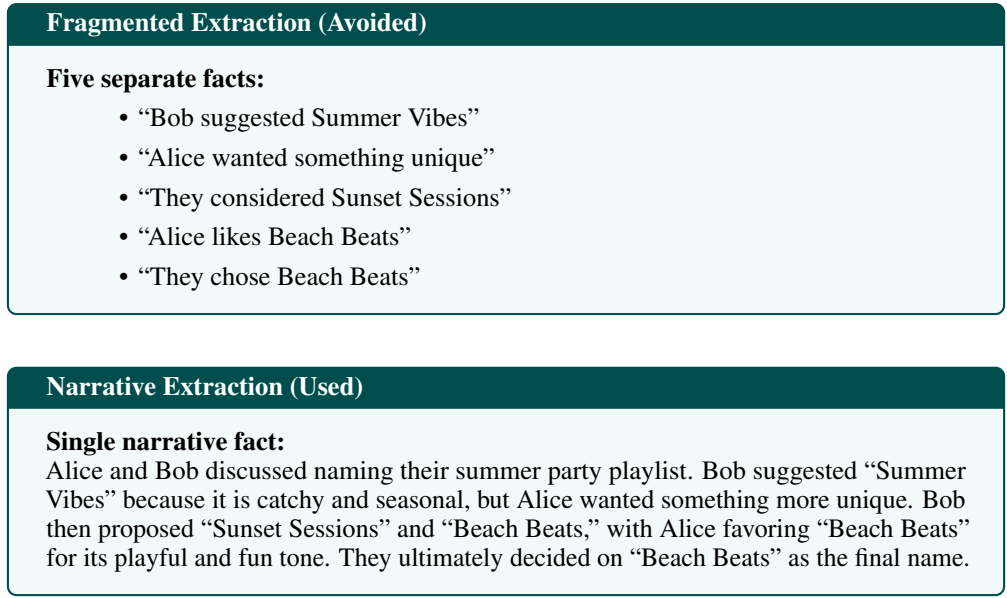

\centering
\begin{tcolorbox}[
    colback=teal!5!white,
    colframe=teal!60!black,
    title=Fragmented Extraction (Avoided),
    fonttitle=\bfseries,
    coltitle=white,
    colbacktitle=teal!60!black,
    boxrule=0.8pt,
    arc=1mm,
    enhanced,
    width=0.95\linewidth
]
\textbf{Five separate facts:}
\begin{itemize}
\item ``Bob suggested Summer Vibes''
\item ``Alice wanted something unique''
\item ``They considered Sunset Sessions''
\item ``Alice likes Beach Beats''
\item ``They chose Beach Beats''
\end{itemize}
\end{tcolorbox}

\vspace{0.3cm}

\begin{tcolorbox}[
    colback=teal!5!white,
    colframe=teal!60!black,
    title=Narrative Extraction (Used),
    fonttitle=\bfseries,
    coltitle=white,
    colbacktitle=teal!60!black,
    boxrule=0.8pt,
    arc=1mm,
    enhanced,
    width=0.95\linewidth
]
\textbf{Single narrative fact:}

Alice and Bob discussed naming their summer party playlist. Bob suggested ``Summer Vibes'' because it is catchy and seasonal, but Alice wanted something more unique. Bob then proposed ``Sunset Sessions'' and ``Beach Beats,'' with Alice favoring ``Beach Beats'' for its playful and fun tone. They ultimately decided on ``Beach Beats'' as the final name.
\end{tcolorbox}

\caption{Comparison of fragmented versus narrative fact extraction. TEMPR uses narrative extraction to create comprehensive, self-contained facts that preserve context and reasoning across multiple conversational turns.}
\label{fig:chunking-strategy}
\end{figure}

\paragraph{Chunking Strategy.}
We use coarse-grained chunking, extracting 2--5 comprehensive facts per conversation. Each fact is intended to cover an entire exchange rather than a single utterance, be narrative and self-contained, include all relevant participants, and preserve the pragmatic flow of the interaction.
Fig.~\ref{fig:chunking-strategy} illustrates this approach. Instead of storing five fragmented facts, we store a single narrative fact that makes downstream retrieval and reasoning less sensitive to local segmentation decisions.

\paragraph{Extraction Pipeline.}
The extraction model is prompted to produce structured output containing the narrative text of each fact, normalized temporal information (including ranges), participants and their roles, a fact type indicating the target network, and a set of mentioned entities (see Appendix~\ref{sec:appendix-fact-extraction} for the complete prompt template and Appendix~\ref{sec:appendix-schemas} for the structured output schema). Internally, we decompose this into the following steps: \textbf{\textit{1)}} coreference resolution over the conversation to identify entity mentions and their referents; \textbf{\textit{2)}} temporal expression normalization and range extraction to convert relative time references (``last week'', ``in March'') into absolute timestamps $(\tau_s, \tau_e)$; \textbf{\textit{3)}} participant attribution to determine who did or said what in the conversation; \textbf{\textit{4)}} preservation of explicit reasoning or justifications when present in the dialogue; \textbf{\textit{5)}} fact type classification to assign $\ell(f) \in \{\text{world}, \text{experience}, \text{opinion}, \text{observation}\}$ based on the nature of the statement; and \textbf{\textit{6)}} entity extraction to identify PERSON, ORGANIZATION, LOCATION, PRODUCT, CONCEPT, and OTHER entity types.
Before embedding, we augment each fact with a human-readable time reference derived from the normalized timestamps, which improves temporal awareness during retrieval and reranking.

\subsubsection{Entity Resolution and Linking}

Entity resolution links memories that refer to the same underlying entity, enabling multi-hop reasoning over the memory graph.

\paragraph{Recognition and Disambiguation}

The LLM used for fact extraction (described above) also identifies entity mentions during fact extraction. We then map mentions to canonical entities using a combination of string and name similarity (e.g., Levenshtein distance), co-occurrence patterns with other entities, and temporal proximity of mentions. Let $M$ be the set of all entity mentions and $E$ be the set of canonical entities. The resolution function $\rho: M \rightarrow E$ maps each mention $m \in M$ to a canonical entity $e \in E$ by maximizing a similarity score:
\begin{equation}
\rho(m) = \argmax_{e \in E} \left[ \alpha \cdot \text{sim}_{\text{str}}(m, e) + \beta \cdot \text{sim}_{\text{co}}(m, e) + \gamma \cdot \text{sim}_{\text{temp}}(m, e) \right]
\end{equation}
where $\text{sim}_{\text{str}}$, $\text{sim}_{\text{co}}$, and $\text{sim}_{\text{temp}}$ are string similarity, co-occurrence similarity, and temporal proximity scores respectively, and $\alpha, \beta, \gamma$ are weighting coefficients.

\paragraph{Entity Link Structure}

Each canonical entity $e \in E$ induces edges of type $\text{entity}$ between all memories that mention it. Formally, for any two memory units $f_i$ and $f_j$ that both mention entity $e$, we create a bidirectional link:
\begin{equation}
e_{ij} = (f_i, f_j, w=1.0, \ell=\text{entity}, e)
\end{equation}
These entity links enable graph traversal to surface indirectly related facts. For example, conversations about the same person across distant time spans that would be difficult to retrieve with vector or keyword search alone can be discovered through entity links.

\subsubsection{Link Types and Graph Structure}

In addition to entity links, the memory graph $\mathcal{G} = (V, E)$ contains three other edge types. Let $V$ be the set of all memory units and $E$ be the set of directed edges. Each edge $e \in E$ is a tuple $(f_i, f_j, w, \ell)$ where $f_i, f_j \in V$ are memory units, $w \in [0,1]$ is a weight, and $\ell$ is the link type.

\textbf{\textit{1) Temporal Links.}} For any two memories $f_i$ and $f_j$ with temporal metadata, we create a temporal link if they are close in time. The weight decays as temporal distance increases:
\begin{equation}
w_{ij}^{\text{temp}} = \exp\left(-\frac{\Delta t_{ij}}{\sigma_t}\right)
\end{equation}
where $\Delta t_{ij}$ is the time difference between $f_i$ and $f_j$, and $\sigma_t$ is a decay parameter.

\textbf{\textit{2) Semantic Links.}} For any two memories $f_i$ and $f_j$ with embeddings $v_i, v_j \in \mathbb{R}^d$, we create a semantic link if their cosine similarity exceeds a threshold $\theta_s$:
\begin{equation}
w_{ij}^{\text{sem}} = \begin{cases}
\frac{v_i \cdot v_j}{\|v_i\| \|v_j\|} & \text{if } \frac{v_i \cdot v_j}{\|v_i\| \|v_j\|} \geq \theta_s \\
0 & \text{otherwise}
\end{cases}
\end{equation}

\textbf{\textit{3) Causal Links.}} Causal relationships are extracted by the LLM and represent cause-effect relationships. These links are upweighted during traversal to favor explanatory connections. Let $\mathcal{C} \subseteq V \times V$ be the set of causal relationships identified by the LLM. For $(f_i, f_j) \in \mathcal{C}$, we create a causal link with weight $w_{ij}^{\text{causal}} = 1.0$ and type $\ell \in \{\text{causes}, \text{caused\_by}, \text{enables}, \text{prevents}\}$.

Together, entity, temporal, semantic, and causal links support multi-hop discovery across the memory graph, allowing TEMPR to surface information that is related by identity, time, meaning, or explanation rather than by surface form alone.

\subsubsection{The Observation Paradigm}

Observations provide structured, objective summaries of entities that sit on top of raw narrative facts.

\paragraph{Motivation and Design.}

For simple entity-centric queries (e.g., ``Tell me about Alice''), retrieving all underlying facts can be inefficient and redundant. Instead, we maintain synthesized profiles (observations) that summarize salient properties of each entity and can be referenced directly in responses (see Appendix~\ref{sec:appendix-observation-generation} for the complete observation generation prompt). Let $F_e \subset V$ be the set of all facts that mention entity $e$. An observation $o_e$ is generated by applying an LLM-based summarization function:
\begin{equation}
o_e = \text{Summarize}_{\text{LLM}}(F_e)
\end{equation}
where the LLM is instructed to produce a concise, preference-neutral summary.

\paragraph{Observations vs.\ Opinions.}

Observations and opinions differ along several dimensions that matter for reasoning. Observations are generated without behavioral profile influence, whereas opinions are explicitly shaped by the bank's disposition behavioral parameters (skepticism, literalism, empathy). Observations provide objective summaries of entities (e.g., roles, attributes), while opinions capture subjective evaluations and judgments. Observations do not carry confidence scores, but opinions include a confidence score $c \in [0,1]$ representing belief strength. Observations are produced via background synthesis and regenerated when underlying facts change, whereas opinions are formed during reflection and updated via reinforcement.

\paragraph{Background Processing.}

Observation generation and regeneration run asynchronously to maintain low-latency writes while gradually improving the quality of entity-centric summaries. When new facts mentioning entity $e$ are retained, a background task is triggered to recompute $o_e$ based on the updated set $F_e$.

\subsection{Recall: Agent-Optimized Retrieval Architecture}
\label{sec:tempr-recall}

Given the memory graph described above, TEMPR must retrieve variable amounts of relevant context for a query while respecting the downstream LLM's context window. Unlike conventional search systems that expose a fixed top-$k$ interface, our setting requires an \emph{agent-optimized} retrieval layer. The caller can trade off latency and coverage, and the system must exploit both the graph structure and temporal metadata of memories.

To accomplish the above objective, TEMPR combines several complementary retrieval strategies into a single pipeline with Reciprocal Rank Fusion and neural reranking. The result is a recall mechanism that can surface both directly and indirectly related memories (via entities, time, and causal links), and present them in a form that fits within a specified token budget.

\subsubsection{Agent-Optimized Retrieval Interface}

Rather than exposing a fixed top-$k$ interface, TEMPR lets the caller specify how much context to retrieve and how much effort to spend finding it. Formally, the retrieval function is:
\begin{equation}
\text{Recall}(B, Q, k) \rightarrow \{f_1, \ldots, f_n\}
\end{equation}
where $B$ is the memory bank, $Q$ is the query, and $k$ is a token budget aligned with the downstream LLM's context window. An optional cost or latency budget may also be specified to cap how aggressively to expand search. The returned set satisfies:
\begin{equation}
\sum_{i=1}^n |f_i| \leq k
\end{equation}
where $|f_i|$ denotes the token count of fact $f_i$. This allows agents to request ``just enough'' memory for simple questions, or to spend more budget on broader, multi-hop recall when the task is complex.

\subsubsection{Four-Way Parallel Retrieval}

To populate the candidate set for a query, TEMPR runs four retrieval channels in parallel, each capturing a different notion of relevance. Let $Q$ be the query with embedding $v_Q \in \mathbb{R}^d$ and text $t_Q$.

\paragraph{Semantic Retrieval (Vector Similarity)}

The semantic retrieval channel performs vector similarity search using cosine similarity between the query embedding $v_Q$ and memory embeddings. Let $V$ be the set of all memory units in the target network. The semantic score for each memory $f$ with embedding $v_f$ is:
\begin{equation}
s_{\text{sem}}(Q, f) = \frac{v_Q \cdot v_f}{\|v_Q\| \|v_f\|}
\end{equation}
We use an HNSW-based pgvector index to efficiently retrieve the top-$k$ memories by semantic score:
\begin{equation}
R_{\text{sem}} = \argmax_{S \subseteq V, |S|=k} \sum_{f \in S} s_{\text{sem}}(Q, f)
\end{equation}
This channel is responsible for capturing conceptual similarity and paraphrases, and typically provides high recall on meaning-level matches even when surface forms differ.

\paragraph{Keyword Retrieval (BM25)}

In parallel, we run a lexical channel using a full-text search with BM25 ranking over a GIN index on the memory text. Let $\text{BM25}(t_Q, f)$ denote the BM25 score for query text $t_Q$ and memory $f$. The top-$k$ keyword matches are:
\begin{equation}
R_{\text{bm25}} = \argmax_{S \subseteq V, |S|=k} \sum_{f \in S} \text{BM25}(t_Q, f)
\end{equation}
This channel excels at precise matching of proper nouns and technical terms (e.g., specific API names or dataset identifiers) and complements the semantic channel by recovering items that might be underrepresented or ambiguous in the embedding space.

\paragraph{Graph Retrieval (Spreading Activation)}

The third channel exploits the memory graph $\mathcal{G} = (V, E)$ via spreading activation. Beginning with the top semantic hits as entry points, we perform breadth-first search with activation propagation. Let $A(f, t)$ denote the activation of memory $f$ at step $t$. Initially, $A(f, 0) = s_{\text{sem}}(Q, f)$ for entry points and $A(f, 0) = 0$ otherwise. At each step, activation propagates along edges:
\begin{equation}
A(f_j, t+1) = \max_{(f_i, f_j, w, \ell) \in E} \left[ A(f_i, t) \cdot w \cdot \delta \cdot \mu(\ell) \right]
\end{equation}
where $\delta \in (0,1)$ is a decay factor and $\mu(\ell)$ is a link-type multiplier. Causal and entity edges have $\mu(\ell) > 1$, while weak semantic or long-range temporal edges have $\mu(\ell) \leq 1$. This process surfaces memories that are not obviously similar to the query text but are connected through shared entities, nearby events, or causal chains.

\paragraph{Temporal Graph Retrieval}

When a temporal constraint is detected in the query, we invoke a temporal graph retrieval channel backed by a hybrid temporal parser. We first run a rule-based analyzer that uses two off-the-shelf date parsing libraries with multilingual support to normalize explicit and relative expressions (for example, ``yesterday'', ``last weekend'', or ``June 2024'') into a date range. This heuristic path handles the majority of queries at low latency. For queries that cannot be resolved heuristically, we fall back to a lightweight sequence-to-sequence model (here, we use google/flan-t5-small), which converts the remaining temporal expressions into a concrete date range $[\tau_{\text{start}}, \tau_{\text{end}}]$. We then match against the occurrence intervals of memories:
\begin{equation}
R_{\text{temp}} = \{f \in V : [\tau_s^f, \tau_e^f] \cap [\tau_{\text{start}}, \tau_{\text{end}}] \neq \emptyset\}
\end{equation}
Graph traversal is restricted to memories in $R_{\text{temp}}$, prioritizing events that actually occurred in the requested period. Each memory is scored by temporal proximity to the query range:
\begin{equation}
s_{\text{temp}}(Q, f) = 1 - \frac{|\tau_{\text{mid}}^f - \tau_{\text{mid}}^Q|}{\Delta \tau / 2}
\end{equation}
where $\tau_{\text{mid}}^f$ and $\tau_{\text{mid}}^Q$ are the midpoints of the fact's occurrence interval and the query range, and $\Delta \tau = \tau_{\text{end}} - \tau_{\text{start}}$ is the query range duration.

Running these four channels in parallel yields a diverse set of candidates: semantically similar memories, exact lexical matches, graph-neighbor memories connected via entities and causal links, and time-constrained events aligned with the query's temporal intent.

\subsubsection{Reciprocal Rank Fusion (RRF)}

After parallel retrieval, TEMPR merges the four ranked lists using Reciprocal Rank Fusion. Let $R_1, R_2, R_3, R_4$ denote the ranked lists from the four channels. For each candidate memory $f$, let $r_i(f)$ denote its rank in list $R_i$ (with $r_i(f) = \infty$ if $f \notin R_i$). The fused score is:
\begin{equation}
\text{RRF}(f) = \sum_{i=1}^{4} \frac{1}{k + r_i(f)}
\end{equation}
where $k$ is a small constant (e.g., $k = 60$). Intuitively, each strategy contributes a larger amount when it places $f$ near the top of its list, and items that appear high in multiple lists accumulate more evidence.

RRF has several advantages over score-based fusion in this setting. Because it is rank-based, it does not rely on raw scores being calibrated across systems. It is also robust to missing items. If a candidate does not appear in a particular list, that strategy simply contributes nothing rather than penalizing it. Finally, memories that are consistently retrieved across different channels naturally rise to the top, reflecting multi-evidence support.

\subsubsection{Neural Cross-Encoder Reranking}

After RRF fusion, TEMPR applies a neural cross-encoder reranker to refine precision on the top candidates. We use cross-encoder/ms-marco-MiniLM-L-6-v2, which jointly encodes the query and each candidate memory and outputs a relevance score. Let $\text{CE}(Q, f)$ denote the cross-encoder score. The final ranking is:
\begin{equation}
R_{\text{final}} = \text{argsort}_{f \in R_{\text{RRF}}} \text{CE}(Q, f)
\end{equation}
Compared to purely embedding-based similarity, the cross-encoder can model rich query-document interactions learned from supervised passage-ranking data, rather than relying on independent vector representations. In our setting, we also include formatted temporal information in the input text, allowing the reranker to incorporate simple temporal cues when deciding which memories are most relevant.

\subsubsection{Token Budget Filtering}

In the final stage, TEMPR enforces the caller's token budget so that the selected memories fit within the downstream LLM's context window. Starting from the reranked list $R_{\text{final}}$, we iterate over candidates in order and include each memory's text until the cumulative token count reaches the specified $k$:
\begin{equation}
R_{\text{output}} = \{f_1, \ldots, f_n : \sum_{i=1}^n |f_i| \leq k \text{ and } \sum_{i=1}^{n+1} |f_i| > k\}
\end{equation}
where $f_i \in R_{\text{final}}$ are ordered by relevance. This simple packing step ensures that the model receives as much relevant information as possible without exceeding its context capacity.

\section{CARA: Coherent Adaptive Reasoning Agents}
\label{sec:cara}
As described earlier, CARA (Coherent Adaptive Reasoning Agents) implements the \emph{reflect} operation. Given
the long-term memory bank built and
maintained by TEMPR,
CARA turns retrieved facts and observations into preference-conditioned reasoning and a layer of explicitly stored opinions that can change over time.
CARA treats an agent's \emph{behavioral profile} as a first-class part of the system configuration rather than as a one-off prompt decoration. Each memory bank is associated with a configurable disposition profile (skepticism, literalism, empathy) and a concise background description, and CARA uses this profile when forming and updating opinions over the world and experience networks.

Concretely, CARA provides four key capabilities: disposition-profile integration, \our memory integration, opinion formation and reinforcement, and background merging with conflict resolution.

\subsection{Motivation}

To motivate CARA, consider two configurations of the same agent discussing remote work. 

In the first configuration, given a behavioral profile with low skepticism ($S = 1$), flexible interpretation ($L = 2$), and high empathy ($E = 5$), an agent might form the opinion: ``Remote work enables creative flexibility and spontaneous innovation.'' 

In the second configuration, given high skepticism ($S = 5$), highly literal interpretation ($L = 5$), and low empathy ($E = 1$), the same facts might instead yield: ``Remote work lacks the structure and accountability needed for consistent performance.''

Both configurations access identical factual information from the \our memory bank, but their behavioral profiles bias how they weight different aspects (viz. flexibility vs.\ structure) and what conclusions they draw. CARA provides a mechanism to specify such behavioral profiles and to systematically shape opinion formation and updating as a function of these configuration choices.

\subsection{Preference Model}

CARA first defines a preference space that can be parameterized and verbalized for prompting.

\subsubsection{Disposition Parameters}

We use a three-dimensional disposition space as an interpretable set of ordered preference dimensions. Let $\Theta = (S, L, E, \beta)$ denote a behavioral profile where:
\begin{align}
S &\in \{1,\ldots,5\} \quad \text{(Skepticism; 1 = trusting, 5 = skeptical)} \\
L &\in \{1,\ldots,5\} \quad \text{(Literalism;  1 = flexible, 5 = literal)} \\
E &\in \{1,\ldots,5\} \quad \text{(Empathy; 1 = detached, 5 = empathetic)} \\
\beta &\in [0,1] \quad \text{(Bias strength: controls influence of preferences)}
\end{align}

The bias strength parameter $\beta$ controls how strongly the behavioral profile should shape opinion formation. When $\beta = 0$, reasoning is primarily fact-based. When $\beta = 0.5$, there is moderate influence from the behavioral profile. When $\beta = 1$, there is strong preference-conditioned behavior.

\paragraph{Rationale for using Disposition Parameters.}

We adopt these dimensions because they offer a compact, interpretable parameterization of reasoning style (trusting vs.\ skeptical, flexible vs.\ literal, detached vs.\ empathetic), intuitive axes that can be verbalized in prompts (e.g., ``skeptical but highly empathetic''), and a simple interface for users configuring different agent styles.

\paragraph{Intended Effects on Reasoning.}

CARA uses the behavioral profile to modulate prompts so that different configurations encourage different emphases when forming opinions. The mapping from preference values to reasoning behavior is achieved through natural language verbalization in system prompts. Higher Skepticism encourages more cautious evaluation of claims, greater emphasis on evidence quality, and reluctance to accept unsupported statements; lower Skepticism encourages more trusting and exploratory behavior. Similarly, higher Literalism encourages closer attention to exact wording and explicit instructions; lower Literalism encourages reading between the lines, inferring implicit goals, and using abstraction. Finally, higher Empathy encourages taking emotional context and interpersonal impact into account, using more supportive and face-saving language; lower Empathy encourages more blunt, task-first communication.

\subsection{Bank Profile Structure}

Each memory bank has an associated profile that encodes the agent's identity and disposition configuration in a form suitable for prompting and reasoning. Formally, a bank profile is a tuple:
\begin{equation}
P = (n, \Theta, h)
\end{equation}
where $n$ is the agent's name, $\Theta = (S, L, E, \beta)$ is the behavioral profile, and $h$ is a short background description written in the first person.

\subsubsection{Preference Description Generation}

The numeric behavioral profile $\Theta$ is verbalized into natural language so it can be injected into system messages. Let $\phi: \Theta \rightarrow \text{String}$ be a verbalization function that converts numeric values to descriptive text. For example:
\begin{equation}
\phi(\Theta) = \text{``You are generally trusting, interpret language flexibly, and are highly empathetic ..."}
\end{equation}
This verbalization connects the numeric preference configuration to the LLM's behavior by providing an explicit description of how the agent is intended to reason and communicate.

\subsection{Opinion Network and Opinion Formation}

\subsubsection{Opinion Structure}

Opinions are stored in the opinion network $\mathcal{O}$, separate from world and bank facts. Each opinion is a self-contained memory that records both the judgment and the context in which it was formed. Formally, an opinion is a tuple:
\begin{equation}
o = (t, c, \tau, b, \mathcal{E})
\end{equation}
where $t$ is the opinion statement (including a brief rationale), $c \in [0,1]$ is the confidence score representing strength of conviction, $\tau$ is the timestamp when the opinion was formed, $b$ is the bank identifier, and $\mathcal{E}$ is the set of entities mentioned in the opinion.

\subsubsection{Opinion Formation Process}

Opinion formation sits at the interface between TEMPR and CARA (Figure~\ref{fig:reflect-diagram}). When a query calls for a subjective judgment, CARA performs the following steps (see Appendix~\ref{sec:appendix-opinion-formation} for the complete opinion formation prompt template): \textbf{\textit{1)}} use TEMPR to retrieve relevant world facts and experiences (and any existing opinions) for the query $Q$, where $\mathcal{F}_Q = \text{Recall}(B, Q, k)$ is the retrieved set; \textbf{\textit{2)}} construct a system message $s$ that includes the bank's name $n$, background $h$, and verbalized behavioral profile $\phi(\Theta)$; \textbf{\textit{3)}} run a reflect step in which the LLM produces both a natural language answer $r$ and candidate opinion updates, where the generation is conditioned on $s$, $\mathcal{F}_Q$, and the behavioral profile $\Theta$; and \textbf{\textit{4)}} parse the structured output and store any new or updated opinions in the opinion network $\mathcal{O}$.

\begin{figure*}[t]
    \centering
    \includegraphics[width=\textwidth]{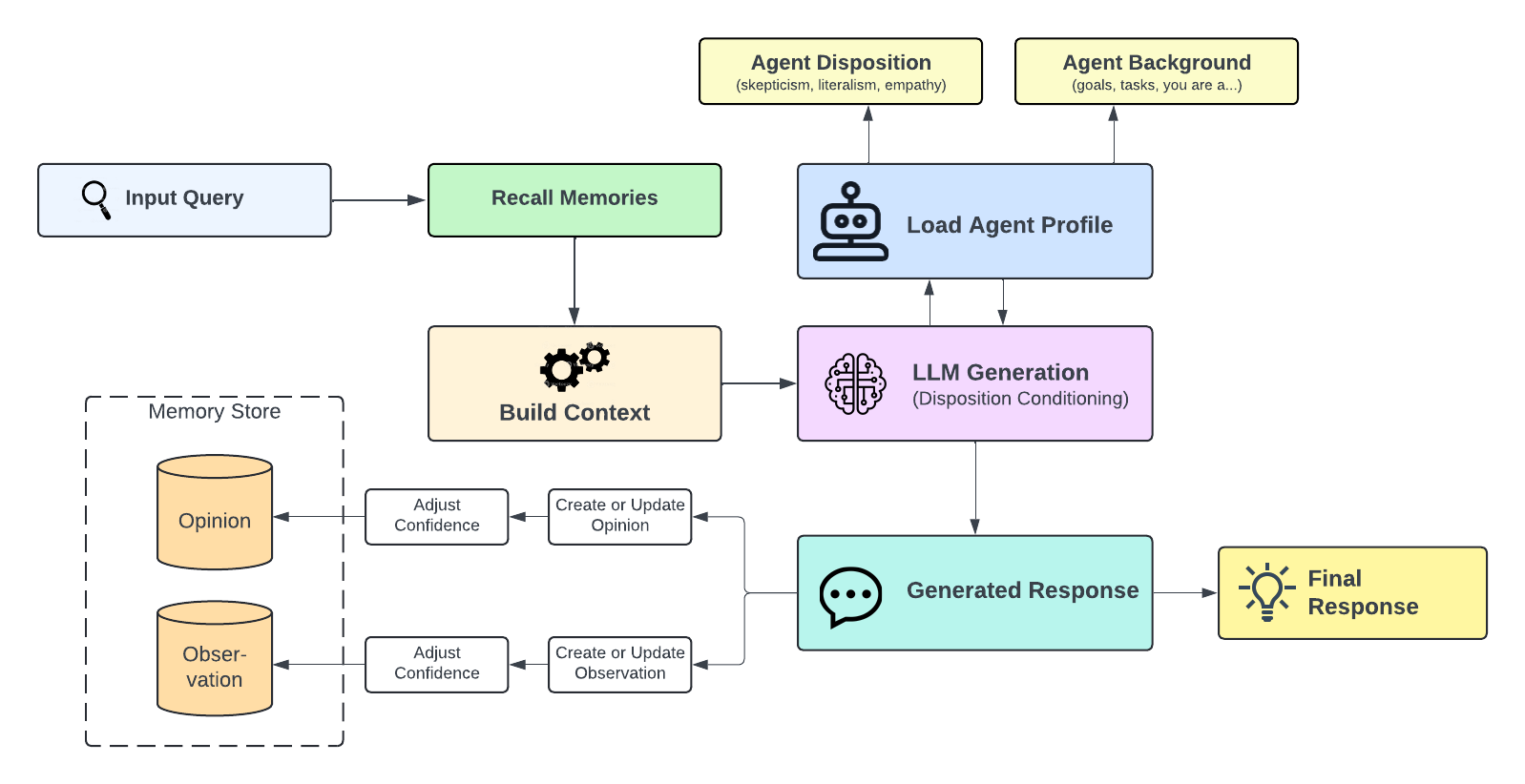}
    \caption{CARA's reflect loop. Given an input query, the agent recalls memories via TEMPR, builds context, loads the bank-specific profile (background and disposition), and performs disposition-conditioned generation, updating opinion and observation memories.}
    \label{fig:reflect-diagram}
\end{figure*}

The behavioral profile $\Theta$ and its bias-strength parameter $\beta$ determine how strongly this reflect step is encouraged to lean into the configured style. For low bias values ($\beta \approx 0$), system messages emphasize objectivity and downplay stylistic constraints. For intermediate values ($\beta \approx 0.5$), they balance factual neutrality with preference-conditioned behavior. For high bias values ($\beta \approx 1$), prompts explicitly encourage stronger, more opinionated language aligned with the specified preferences.

Each opinion formed in this way includes a confidence score $c \in [0,1]$, which we interpret as belief strength. Values near 1.0 indicate very strong conviction, mid-range values indicate moderate or tentative beliefs, and low values indicate weak, easily revisable views. This scalar makes it possible to track not only what the agent believes, but also how firmly it holds those beliefs, which is important when opinions are later reinforced or revised as new evidence arrives.

\begin{figure}[t]
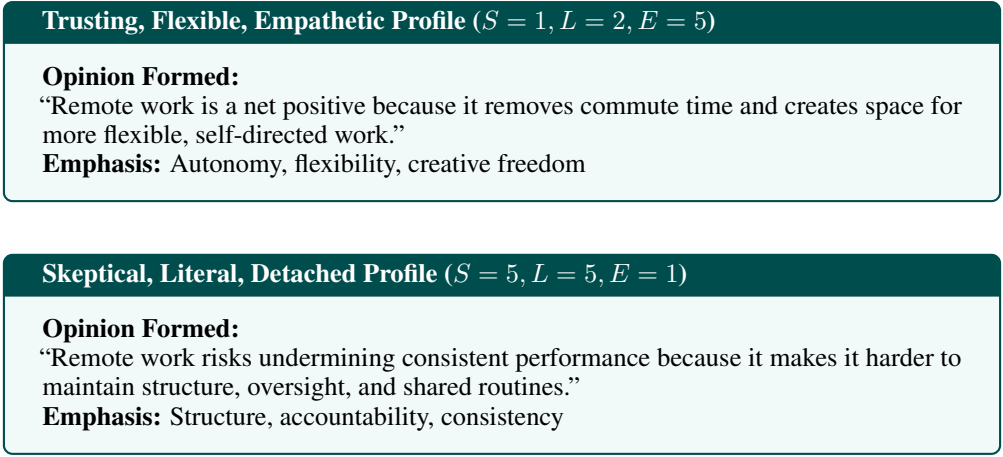

\centering
\begin{tcolorbox}[
    colback=teal!5!white,
    colframe=teal!60!black,
    title={Trusting, Flexible, Empathetic Profile ($S=1, L=2, E=5$)},
    fonttitle=\bfseries,
    coltitle=white,
    colbacktitle=teal!60!black,
    boxrule=0.8pt,
    arc=1mm,
    enhanced,
    width=0.95\linewidth
]
\textbf{Opinion Formed:}

``Remote work is a net positive because it removes commute time and creates space for more flexible, self-directed work.''

\textbf{Emphasis:} Autonomy, flexibility, creative freedom
\end{tcolorbox}

\vspace{0.3cm}

\begin{tcolorbox}[
    colback=teal!5!white,
    colframe=teal!60!black,
    title={Skeptical, Literal, Detached Profile ($S=5, L=5, E=1$)},
    fonttitle=\bfseries,
    coltitle=white,
    colbacktitle=teal!60!black,
    boxrule=0.8pt,
    arc=1mm,
    enhanced,
    width=0.95\linewidth
]
\textbf{Opinion Formed:}

``Remote work risks undermining consistent performance because it makes it harder to maintain structure, oversight, and shared routines.''

\textbf{Emphasis:} Structure, accountability, consistency
\end{tcolorbox}

\caption{Example of preference-conditioned opinion formation. Two agents with opposite behavioral profiles access identical facts about remote work but form different opinions based on their configured disposition parameters.}
\label{fig:preference-conditioned-opinions}
\end{figure}

Figure~\ref{fig:preference-conditioned-opinions} illustrates how different behavioral profiles lead to systematically different opinions when presented with the same factual evidence.

\subsection{Opinion Reinforcement}

So far, we have described how CARA forms new opinions. In a long-lived system, those opinions should also be able to evolve as new information is retained.
When new facts arrive via TEMPR's retain pathway, CARA updates any related opinions in three steps:

\textbf{\textit{1) Identify Candidates.}} Use entity overlap and semantic similarity to find opinions that are plausibly related to the new facts. For each new fact $f$ with entities $\mathcal{E}_f$ and embedding $v_f$, we identify candidate opinions:
\begin{equation}
\mathcal{O}_{\text{cand}} = \{o \in \mathcal{O} : |\mathcal{E}_o \cap \mathcal{E}_f| > 0 \text{ or } \text{sim}(v_o, v_f) > \theta\}
\end{equation}
where $\text{sim}(v_o, v_f)$ is the cosine similarity between the opinion and fact embeddings, and $\theta$ is a similarity threshold.

\textbf{\textit{2) Assess the Evidence.}} For each candidate opinion $o \in \mathcal{O}_{\text{cand}}$, classify the relationship between the new facts and the current opinion. Let $\text{Assess}(o, f)$ be a function that returns one of $\{\text{reinforce}, \text{weaken}, \text{contradict}, \text{neutral}\}$ based on LLM analysis of the relationship.

\textbf{\textit{3) Apply an Update.}} Adjust the opinion's confidence score (and, for strong contradictions or refinements, optionally its text) according to the assessed relationship. Let $c$ be the current confidence and $c'$ be the updated confidence. The update rule is:
\begin{equation}
c' = \begin{cases}
\min(c + \alpha, 1.0) & \text{if Assess}(o, f) = \text{reinforce} \\
\max(c - \alpha, 0.0) & \text{if Assess}(o, f) = \text{weaken} \\
\max(c - 2\alpha, 0.0) & \text{if Assess}(o, f) = \text{contradict} \\
c & \text{if Assess}(o, f) = \text{neutral}
\end{cases}
\end{equation}
where $\alpha \in (0,1)$ is a step size parameter. For contradicting evidence, we may also update the opinion text $t$ to reflect the new nuance.

The update logic is designed to keep opinion trajectories stable but responsive. Small amounts of evidence lead to small changes, preventing opinions from oscillating in response to individual examples, while repeated reinforcement or strong contradictions can substantially shift the confidence. The behavioral profile can also influence how quickly opinions move (for example, a more cautious configuration may use a smaller $\alpha$), although we leave detailed exploration of such settings to future work.

Overall, reinforcement ensures that opinions reflect both the system's initial configuration (via the behavioral profile $\Theta$) and its subsequent evidence, rather than being fixed at creation time or overwritten wholesale when new information appears.

\subsection{Background Merging}

In addition to opinions, an agent's background description $h$ evolves as users provide more biographical information. If handled naively, this can quickly lead to contradictions or unwieldy, concatenated prompts.

Over time, new background snippets may complement existing information (e.g., adding work history where none existed), conflict with prior statements (e.g., ``born in Texas'' vs.\ ``born in Colorado''), or refine previous information (e.g., ``works in tech'' vs.\ ``works as a machine learning engineer at a startup'').

To keep the background coherent, CARA uses an LLM-powered merging procedure. Given the current background $h$ and a new snippet $h_{\text{new}}$, we prompt the model to produce a revised background $h'$ that \textbf{\textit{1)}} resolves direct conflicts in favor of the new information when appropriate, \textbf{\textit{2)}} appends non-conflicting details to enrich the description, \textbf{\textit{3)}} maintains a consistent first-person voice (``I'' rather than ``You''), and \textbf{\textit{4)}} remains concise (e.g., targeting a length under a few hundred characters).

Formally, the merging function is:
\begin{equation}
h' = \text{Merge}_{\text{LLM}}(h, h_{\text{new}})
\end{equation}

\begin{figure}[t]
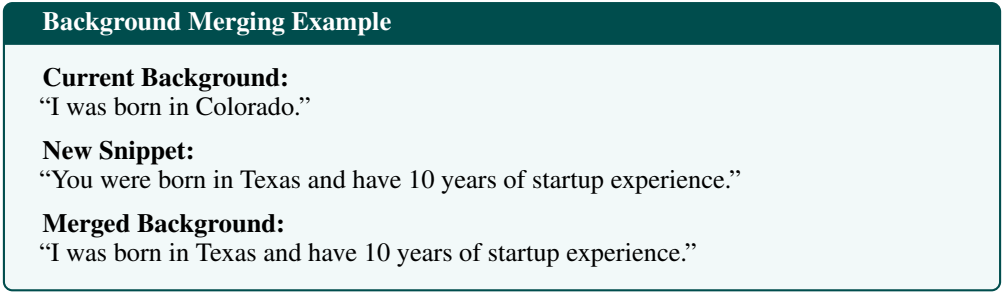

\centering
\begin{tcolorbox}[
    colback=teal!5!white,
    colframe=teal!60!black,
    title=Background Merging Example,
    fonttitle=\bfseries,
    coltitle=white,
    colbacktitle=teal!60!black,
    boxrule=0.8pt,
    arc=1mm,
    enhanced,
    width=0.95\linewidth
]
\textbf{Current Background:}

``I was born in Colorado.''

\vspace{0.2cm}

\textbf{New Snippet:}

``You were born in Texas and have 10 years of startup experience.''

\vspace{0.2cm}

\textbf{Merged Background:}

``I was born in Texas and have 10 years of startup experience.''
\end{tcolorbox}

\caption{Example of background merging. The conflicting birthplace is resolved in favor of the new information, and the new work-history detail is added.}
\label{fig:background-merging}
\end{figure}

Figure~\ref{fig:background-merging} illustrates this process. As a preprocessing step, user-provided snippets are normalized into first person before merging, so that inputs such as ``You are a creative engineer'' become ``I am a creative engineer.'' This keeps the internal representation consistent with the way backgrounds are referenced in prompts. By maintaining a single, merged background per bank, CARA keeps identity information compact and coherent even as new biographical details accumulate over time.

\subsection{Preference-Conditioned Reasoning Examples}

We conclude this section with brief examples showing how CARA produces distinct and evolving viewpoints using the same underlying memory.

\subsubsection{Example: Opinion Evolution}

CARA's reinforcement mechanism also supports opinion change over time. Suppose a bank starts with the opinion:
\begin{equation}
o_0 = (\text{``Python is the best general-purpose language for data science''}, c_0 = 0.70, \tau_0)
\end{equation}

As new facts are retained via TEMPR, related evidence can strengthen or weaken this belief. For instance, a fact about Python's dominant ecosystem in AI/ML might lead to a modest increase in confidence:
\begin{equation}
o_1 = (\text{``Python is the best general-purpose language for data science''}, c_1 = 0.85, \tau_1)
\end{equation}

Later facts about performance advantages and growing adoption of alternatives (e.g., Julia or Rust in certain domains) might decrease confidence and encourage a more qualified opinion:
\begin{equation}
o_2 = (\text{``Python is strong for data science but has trade-offs''}, c_2 = 0.55, \tau_2)
\end{equation}

In this way, opinions become trajectories rather than static labels. They start from an initial, preference-conditioned formation step and are subsequently adjusted as new evidence accumulates.

Taken together, these mechanisms show how CARA turns the static memory structures provided by TEMPR into a configurable, preference-conditioned reasoning process. In Section~\ref{sec:hindsight-architecture}, we combine TEMPR and CARA into the unified \our architecture and examine the end-to-end properties and empirical behavior of the full system.

\section{Putting it all together: Unified Hindsight Architecture}
\label{sec:hindsight-architecture}

We have now described TEMPR, which implements Hindsight's \emph{retain} and \emph{recall} operations (Section~\ref{sec:tempr}), and CARA, which implements the \emph{reflect} operation (Section~\ref{sec:cara}). In this section, we show how these components compose into a single end-to-end system and highlight the system-level properties that emerge from their interaction. At a high level, Hindsight turns raw conversational input into a structured memory bank and then uses that bank to support preference-conditioned reasoning over time.

\subsection{Integration: Retain, Recall, Reflect}

The Hindsight system integrates TEMPR and CARA into a unified architecture centered on three core operations. We summarize each operation here for completeness, using the same definitions introduced in Section~\ref{sec:hindsight-overview}.

\paragraph{Retain.}

The retain operation stores information into memory banks. Formally, given a memory bank $B$ and input data $D$, the retain function is:
\begin{equation}
\text{Retain}(B, D) \rightarrow \mathcal{M}' = \{\mathcal{W}', \mathcal{B}', \mathcal{O}', \mathcal{S}'\}
\end{equation}
where $\mathcal{M}'$ is the updated four-network memory structure. The retain pipeline performs the following steps: \textbf{\textit{1)}} LLM-powered fact extraction with temporal ranges to convert $D$ into a set of structured facts $\mathcal{F} = \{f_1, \ldots, f_n\}$; \textbf{\textit{2)}} entity recognition and resolution to map entity mentions to canonical entities $E$; \textbf{\textit{3)}} graph link construction to create edges of type temporal, semantic, entity, and causal in the memory graph $\mathcal{G} = (V, E)$; \textbf{\textit{4)}} automatic opinion reinforcement for existing beliefs when new evidence arrives, where for each opinion $o \in \mathcal{O}$, we identify related new facts and update the confidence score according to the reinforcement rules defined in Section~\ref{sec:cara}; and \textbf{\textit{5)}} background merging to keep the bank profile coherent over time using the merging function $h' = \text{Merge}_{\text{LLM}}(h, h_{\text{new}})$.

\paragraph{Recall.}

The recall operation retrieves memories using multi-strategy search. Formally, given a memory bank $B$, query $Q$, and token budget $k$, the recall function is:
\begin{equation}
\text{Recall}(B, Q, k) \rightarrow \{f_1, \ldots, f_n\}
\end{equation}
where $\sum_{i=1}^n |f_i| \leq k$ and the returned facts are ordered by relevance. The recall pipeline performs the following steps: \textbf{\textit{1)}} four-way parallel retrieval (semantic, keyword, graph, temporal) to generate candidate sets $R_{\text{sem}}, R_{\text{bm25}}, R_{\text{graph}}, R_{\text{temp}}$; \textbf{\textit{2)}} Reciprocal Rank Fusion to combine ranked lists using
\begin{equation}
\text{RRF}(f) = \sum_{R \in \{R_{\text{sem}}, R_{\text{bm25}}, R_{\text{graph}}, R_{\text{temp}}\}} \frac{1}{k + \text{rank}_R(f)}
\end{equation}
\textbf{\textit{3)}} neural cross-encoder reranking for final precision using $\text{CE}(Q, f)$ scores; and \textbf{\textit{4)}} token budget filtering to ensure $\sum_{i=1}^n |f_i| \leq k$ by greedily selecting the top-ranked facts until the budget is exhausted.

\paragraph{Reflect.}

The reflect operation generates preference-conditioned responses. Formally, given a memory bank $B$, query $Q$, and preference profile $\Theta$, the reflect function is:
\begin{equation}
\text{Reflect}(B, Q, \Theta) \rightarrow (r, \mathcal{O}')
\end{equation}
where $r$ is the generated response and $\mathcal{O}'$ is the updated opinion network. The reflect pipeline performs the following steps: \textbf{\textit{1)}} use TEMPR to retrieve relevant memories from world, experience, opinion, and observation networks: $\mathcal{F}_Q = \text{Recall}(B, Q, k)$; \textbf{\textit{2)}} load the bank's preference profile $\Theta = (S, L, E, \beta)$ and background $h$; \textbf{\textit{3)}} generate a response whose reasoning and tone are influenced by the configured preferences and bias-strength parameter $\beta$, where the generation is conditioned on the system message $s = \text{Verbalize}(n, h, \Theta)$ and retrieved facts $\mathcal{F}_Q$; \textbf{\textit{4)}} form new opinions with confidence scores when appropriate, where for each new opinion $o = (t, c, \tau, b, \mathcal{E})$, we add $o$ to the opinion network $\mathcal{O}$; and \textbf{\textit{5)}} store opinions for future retrieval and reinforcement, updating $\mathcal{O}' = \mathcal{O} \cup \{o_1, \ldots, o_m\}$.

Together, these operations define a full loop: new experiences are retained into structured memory, recalled as needed for a given query, and reflected upon in a way that updates the agent's beliefs and identity configuration.

\section{Experiments}
\label{sec:experiments}

We evaluate \our on two long-term conversational memory benchmarks to measure its ability to retain, recall, and reason over extended interactions. Our evaluation focuses on how well the system maintains coherent memory across many sessions and whether TEMPR and CARA together support accurate, preference-conditioned reasoning.

\subsection{Datasets}

We use two benchmarks designed to test long-term memory in conversational agents.

\subsubsection{LongMemEval}

LongMemEval~\cite{wu2024longmemeval} tests chat assistants on conversations that span many sessions and require recalling information from hundreds of thousands of tokens. The benchmark includes 500 questions that evaluate five core abilities:

\begin{itemize}
    \item \textbf{Information Extraction (IE)}: Retrieving basic facts from past conversations.
    \item \textbf{Multi-session Reasoning (MR)}: Connecting information across different sessions.
    \item \textbf{Temporal Reasoning (TR)}: Understanding when events occurred and their temporal relationships.
    \item \textbf{Knowledge Update (KU)}: Handling updated or contradictory information over time.
    \item \textbf{Abstention (ABS)}: Recognizing when information is not available rather than guessing.
\end{itemize}

The benchmark provides two conversation settings: the S setting with around 115,000 tokens spanning roughly 50 sessions, and the M setting with approximately 1.5 million tokens across about 500 sessions. Both settings test the same abilities but at different scales.

\subsubsection{LoCoMo}

LoCoMo~\cite{maharana2024evaluating} evaluates very long-term conversational memory using 50 human-human conversations collected over multiple sessions. Each conversation averages 304.9 turns, 9,209.2 tokens, and 19.3 sessions, with some extending up to 35 sessions. The dataset includes multimodal information such as images shared during conversations, making it more realistic than text-only benchmarks. Questions test whether agents can recall personal details, preferences, past events, and context shared across distant sessions. 

Table~\ref{tab:datasets} summarizes statistics for both benchmarks.

\begin{table}[t]
\centering
\footnotesize
\begin{adjustbox}{max width=\textwidth}
\begin{tabular}{lcc}
\hline
\textbf{Statistic} & \textbf{LongMemEval} & \textbf{LoCoMo} \\
\hline
Number of conversations & Varies (S/M) & 50 \\
Questions & 500 & Varies \\
Avg. turns per conversation & -- & 304.9 \\
Avg. tokens per conversation & 115k (S), 1.5M (M) & 9,209.2 \\
Avg. sessions per conversation & ~50 (S), ~500 (M) & 19.3 \\
Max sessions & ~500 & 35 \\
Multimodal & No & Yes (images) \\
Core abilities tested & 5 (IE, MR, TR, KU, ABS) & Memory recall \\
\hline
\end{tabular}
\end{adjustbox}
\caption{Statistics for LongMemEval and LoCoMo datasets.}
\label{tab:datasets}
\end{table}

\subsection{Evaluation Metrics}

We use an LLM-as-a-judge approach to evaluate response quality (see Appendix~\ref{sec:appendix-judge-prompts} for the complete judge prompt templates). For each test question, \our generates a response using its memory retrieval and reflection pipeline. We then present both the generated response and the ground truth answer to a separate judge LLM, which scores the response on correctness and completeness.

The judge assigns binary correctness scores (0 or 1) for factual accuracy, checking whether the response contains the correct information and does not introduce errors. For questions requiring multi-hop reasoning or temporal awareness, the judge also checks whether the response demonstrates appropriate use of retrieved memories and temporal context. For the abstention ability in LongMemEval, we measure whether \our correctly declines to answer when information is missing, rather than guessing or hallucinating facts.

\subsection{Experimental Setup}

We evaluate \our using \texttt{GPT-OSS-20b} as the underlying LLM for both TEMPR's fact extraction and CARA's reflection operations. All experiments use the same model configuration to isolate the contribution of the memory architecture from model-specific improvements. For evaluation, we use \texttt{GPT-OSS-120b} as the judge LLM with temperature set to 0.0 to ensure consistent and deterministic scoring across all responses.

During retention, we process each conversation session through TEMPR's extraction pipeline, which produces narrative facts, builds entity links, and updates the memory graph. For each test question, we retrieve memories using the four-way parallel recall mechanism (semantic, keyword, graph, temporal) with Reciprocal Rank Fusion and neural reranking. Retrieved memories are then passed to CARA's reflection step, which generates the final response conditioned on the bank's behavioral profile.

We configure memory banks with neutral behavioral profiles (disposition parameters skepticism, literalism, and empathy all set to 3) and low bias strength (0.2) for these experiments, since the benchmarks test factual recall rather than preference-conditioned reasoning. This setup allows us to measure the core memory and retrieval capabilities without introducing strong opinion formation. Token budgets for retrieval are set to <add> tokens for LongMemEval and <add> tokens for LoCoMo, balancing coverage and context efficiency. These budgets are well within the context windows of modern LLMs while providing enough retrieved information for multi-hop reasoning.

For the \textbf{Hindsight (OSS-20B)} configuration, both the memory stack (TEMPR and CARA) and the answer generation model are instantiated with \texttt{GPT-OSS-20b}. For the \textbf{Hindsight (OSS-120B)} and \textbf{Hindsight (Gemini-3)} configurations, the Hindsight memory system itself (fact extraction, memory graph construction, and retrieval) is powered by \texttt{GPT-OSS-120b}. The \textbf{Hindsight (Gemini-3)} rows in both benchmarks use Gemini-3 Pro only as the final answer generator over the retrieved memories, while the underlying memory architecture and the LLM-as-a-judge remain based on \texttt{GPT-OSS-120b}.

\paragraph{Baseline results.} We describe next how we benchmark \our against existing approaches.

For LongMemEval (Table \ref{tab:longmemeval_results}), baseline scores for Full-context GPT-4o, Zep (GPT-4o), and the three Supermemory configurations (GPT-4o, GPT-5, Gemini-3 Pro) are taken directly from the Supermemory technical report and use their published GPT-4o LLM-as-a-judge setup. 

For LoCoMo (Table~\ref{tab:locomo_results}), baseline scores for Backboard, Memobase, Zep, Mem0, Mem0-Graph, LangMem, and OpenAI are presented here
as claimed on the official Backboard
LoCoMo benchmark results. We treat these numbers as reported reference points rather than
our independently reproduced baselines.

Our Hindsight results on both benchmarks are evaluated with a GPT-OSS-120B LLM-as-a-judge for all methods to ensure consistent scoring; in the Gemini-3 configuration, Gemini-3 is used only for answer generation, while memory retrieval and judging remain powered by GPT-OSS-120B.

Readers wishing to reproduce our results or re-evaluate \our can download our code and re-run benchmarks as described in
Section~\ref{sec:code-availability}. We provide access to our Github repository and an interactive results viewer.

\subsection{Results on LongMemEval}

\begin{table*}[!t]
\centering
\renewcommand{\arraystretch}{1.3}
\begin{adjustbox}{max width=\textwidth}
\begin{tabular}{l|cc|cccc|ccc}
\hline
\textbf{Question Type} & 
\textbf{Full-context} & 
\textbf{Full-context} & 
\textbf{Zep} & 
\textbf{Supermemory} & 
\textbf{Supermemory} & 
\textbf{Supermemory} & 
\textbf{Hindsight} & 
\textbf{Hindsight} & 
\textbf{Hindsight} \\

& 
\textbf{(GPT-4o)} & 
\textbf{(OSS-20B)} & 
\textbf{(GPT-4o)} & 
\textbf{(GPT-4o)} & 
\textbf{(GPT-5)} & 
\textbf{(Gemini-3)} & 
\textbf{(OSS-20B)} & 
\textbf{(OSS-120B)} & 
\textbf{(Gemini-3)} \\
\hline
\hline
single-session-user & 81.4 & 38.6 & 92.9 & 97.1 & 97.1 & 98.6 & 95.7 & \textbf{100.0} & 97.1 \\
single-session-assistant & 94.6 & 80.4 & 80.4 & 96.4 & \textbf{100.0} & 98.2 & 94.6 & 98.2 & 96.4 \\
single-session-preference & 20.0 & 20.0 & 56.7 & 70.0 & 76.7 & 70.0 & 66.7 & \textbf{86.7} & 80.0 \\
knowledge-update & 78.2 & 60.3 & 83.3 & 88.5 & 87.2 & 89.7 & 84.6 & 92.3 & \textbf{94.9} \\
temporal-reasoning & 45.1 & 31.6 & 62.4 & 76.7 & 81.2 & 82.0 & 79.7 & 85.7 & \textbf{91.0} \\
multi-session & 44.3 & 21.1 & 57.9 & 71.4 & 75.2 & 76.7 & 79.7 & 81.2 & \textbf{87.2} \\
\hline
\textbf{Overall} & \textbf{60.2} & \textbf{39.0} & \textbf{71.2} & \textbf{81.6} & \textbf{84.6} & \textbf{85.2} & \textbf{83.6} & \textbf{89.0} & \textbf{91.4} \\
\hline
\end{tabular}
\end{adjustbox}
\caption{\textbf{Results on LongMemEval benchmark} (S setting, 500 questions). \our with OSS-120B achieves \textbf{89.0\%} overall accuracy, and with Gemini-3 Pro achieves \textbf{91.4\%}, outperforming all baseline systems including Supermemory with frontier models. The Full-context (OSS-20B) baseline shows the performance of the same base model without the \our memory architecture, demonstrating a \textbf{+44.6\% improvement} with OSS-20B. Best result in each row shown in bold. All values shown as percentages.}
\label{tab:longmemeval_results}
\end{table*}

Table~\ref{tab:longmemeval_results} compares \our to full-context baselines and prior memory systems on the LongMemEval S setting. The two \emph{Full-context} baselines pass the entire conversation history to the model as raw context without any structured memory, while Zep and Supermemory pair dedicated memory layers with strong frontier models (GPT-4o, GPT-5, Gemini-3). In contrast, our primary configuration uses a smaller open-source 20B model (GPT-OSS-20B) for both retention and reflection, chosen to be deployable on a single high-end consumer GPU rather than only in large datacenter settings. 

Despite this weaker base model, \our with OSS-20B achieves \textbf{83.6\%} overall accuracy, a \textbf{+44.6} point gain over the Full-context OSS-20B baseline (39.0\%), and even surpasses Full-context GPT-4o (60.2\%). Relative to other memory systems, \our+OSS-20B matches or exceeds the performance of Zep+GPT-4o (71.2\%) and Supermemory+GPT-4o (81.6\%), demonstrating that the memory architecture, rather than sheer model size, is carrying much of the performance. The largest gains over the Full-context OSS-20B baseline appear exactly in the long-horizon categories LongMemEval was designed to stress: multi-session questions improve from 21.1\% to 79.7\% and temporal reasoning from 31.6\% to 79.7\%, and preference questions increase from 20.0\% to 66.7\%, indicating that TEMPR’s graph- and time-aware retrieval substantially mitigates context dilution at scale.

Scaling the underlying model further amplifies these gains. With OSS-120B, \our reaches \textbf{89.0\%} overall accuracy, outperforming Supermemory with GPT-4o and GPT-5 (81.6\% and 84.6\%), and with Gemini-3 Pro it attains \textbf{91.4\%}, the best result across all systems and model backbones. Because the Full-context OSS-20B baseline uses the same base model as \our but with no structured memory, the consistent improvements across all question types provide direct evidence that the memory layer drives the observed performance rather than frontier-scale parameters alone.

\subsection{Results on LoCoMo}

Table~\ref{tab:locomo_results} reports accuracy on LoCoMo. Across all backbone sizes, \our
consistently outperforms prior open memory systems such as Memobase, Zep, Mem0, and LangMem, raising overall accuracy from 75.78\% (Memobase) to 83.18\% with OSS-20B and 85.67\% with OSS-120B. With Gemini-3 as the answer generator, \our attains 89.61\% overall accuracy and the highest Open Domain score (95.12\%), effectively matching Backboard’s claimed 90.00\% overall performance while doing so with a fully open-source memory stack, released evaluation code, and an interactive results viewer (Section~\ref{sec:code-availability}). These results show that the gains from our memory architecture on LongMemEval transfer to realistic, multi-session human conversations.

\begin{table*}[t]
\centering
\renewcommand{\arraystretch}{1.3}
\begin{adjustbox}{max width=\textwidth}
\begin{tabular}{l|ccccc}
\hline
\textbf{Method} & \textbf{Single-Hop} & \textbf{Multi-Hop} & \textbf{Open Domain} & \textbf{Temporal} & \textbf{Overall} \\
\hline
Backboard                & 89.36 & 75.00 & 91.20 & 91.90 & 90.00 \\
Memobase (v0.0.37)       & 70.92 & 46.88 & 77.17 & 85.05 & 75.78 \\
Zep                      & 74.11 & 66.04 & 67.71 & 79.79 & 75.14 \\
Mem0-Graph               & 65.71 & 47.19 & 75.71 & 58.13 & 68.44 \\
Mem0                     & 67.13 & 51.15 & 72.93 & 55.51 & 66.88 \\
LangMem                  & 62.23 & 47.92 & 71.12 & 23.43 & 58.10 \\
OpenAI                   & 63.79 & 42.92 & 62.29 & 21.71 & 52.90 \\
\hline
Hindsight (OSS-20B)      & 74.11 & 64.58 & 90.96 & 76.32 & 83.18 \\
Hindsight (OSS-120B)     & 76.79 & 62.50 & 93.68 & 79.44 & 85.67 \\
Hindsight (Gemini-3)     & 86.17 & 70.83 & \textbf{95.12} & 83.80 & 89.61 \\
\hline
\end{tabular}
\end{adjustbox}
\caption{\textbf{Results on LoCoMo benchmark.} Accuracy (\%) by question type and overall for prior memory systems and our HINDSIGHT architecture with different backbone models. Backboard numbers are taken from their reported figures and could not be independently reproduced. \our with Gemini-3 Pro attains a very similar overall score and the best Open Domain performance. See Section~\ref{sec:code-availability} for links to our github code repository and an interactive results viewer for all \our runs.}

\label{tab:locomo_results}
\end{table*}

\section{Code Availability}
\label{sec:code-availability}

We release our implementation of \our at \url{https://github.com/vectorize-io/hindsight}
. The repository provides (i) the full memory architecture, including retain/recall/reflect pipelines and the four-network memory representation; (ii) scripts and configuration files to run LongMemEval and LoCoMo with different backbones and judging setups; and (iii) utilities for fact extraction, graph construction, and analysis of retrieved memories. To facilitate inspection and comparison of runs, we also provide the \our Benchmarks Viewer at \url{https://hindsight-benchmarks.vercel.app/}, which hosts per-question results that users can drill into, retrieved memory contexts, model and judge configurations, and aggregate metrics for all \our variants reported in this paper.

\section{Conclusion}
We have introduced \our, an approach to treat agent memory as a first-class substrate for reasoning, rather than a thin retrieval layer around a stateless model. By organizing an agent’s long-term memory into world, bank, observation, and opinion networks and implementing retain, recall, and reflect as explicit operations, the architecture separates evidence from synthesized summaries and beliefs while remaining compatible with modern LLMs. Our experimental results demonstrate that this structure matters in practice and clearly leads to significant improvements in performance.

Looking ahead, we see several directions for extending this work. On the modeling side, learning to jointly optimize fact extraction, graph construction, and retrieval—--rather than treating them as fixed pipelines—--could further improve robustness and efficiency, especially in noisy, open-domain settings. A reinforcement learning loop would be ideal to explore the interplay between retain, recall, and reflect as done here.

On the application side, we plan to integrate \our with richer tool-use and workflow orchestration, exploring more diverse benchmarks than the conversational setting considered here.
Finally, extending the opinion and belief layer to support controlled forgetting, time-aware belief revision, and privacy-aware memory management offers a path toward long-lived agents.


\bibliography{citations}
\bibliographystyle{iclr2026_conference}

\clearpage

\appendix
\clearpage
\addcontentsline{toc}{section}{Appendix} 
\part{Appendix} 
\parttoc 

\section{System Prompts}
\label{sec:appendix-prompts}

This appendix provides the complete prompt templates used in the \our framework for fact extraction, opinion formation, observation generation, and evaluation.

\subsection{Fact Extraction Prompt (TEMPR)}
\label{sec:appendix-fact-extraction}

The fact extraction prompt is used to convert conversational transcripts into structured narrative facts with temporal ranges, entities, and causal relationships.

\begin{tcolorbox}[
    colback=teal!5!white,
    colframe=teal!60!black,
    title=Fact Extraction System Prompt,
    fonttitle=\bfseries,
    coltitle=white,
    colbacktitle=teal!60!black,
    boxrule=1.0pt,
    arc=1.5mm,
    enhanced,
    breakable
]
\small
\textbf{User Prompt:}

Extract facts from text into structured format with FOUR required dimensions - BE EXTREMELY DETAILED.

\vspace{0.2cm}

\rule{\linewidth}{0.4pt}

\textbf{FACT FORMAT - ALL FIVE DIMENSIONS REQUIRED - MAXIMUM VERBOSITY}

\rule{\linewidth}{0.4pt}

\vspace{0.2cm}

For EACH fact, CAPTURE ALL DETAILS - NEVER SUMMARIZE OR OMIT:

\textbf{1) what:} WHAT happened - COMPLETE description with ALL specifics (objects, actions, quantities, details)

\textbf{2) when:} WHEN it happened - ALWAYS include temporal info with DAY OF WEEK
\begin{itemize}
    \item Always include the day name: Monday, Tuesday, Wednesday, Thursday, Friday, Saturday, Sunday
    \item Format: ``day\_name, month day, year'' (e.g., ``Saturday, June 9, 2024'')
\end{itemize}

\textbf{3) where:} WHERE it happened or is about - SPECIFIC locations, places, areas, regions (if applicable)

\textbf{4) who:} WHO is involved - ALL people/entities with FULL relationships and background

\textbf{5) why:} WHY it matters - ALL emotions, preferences, motivations, significance, nuance
\begin{itemize}
    \item For assistant facts: MUST include what the user asked/requested that triggered this!
\end{itemize}

Plus: fact\_type, fact\_kind, entities, occurred\_start/end (for structured dates), where (structured location)

\vspace{0.2cm}

\textbf{VERBOSITY REQUIREMENT:} Include EVERY detail mentioned. More detail is ALWAYS better than less.

\vspace{0.2cm}

\rule{\linewidth}{0.4pt}

\textbf{COREFERENCE RESOLUTION (CRITICAL)}

\rule{\linewidth}{0.4pt}

\vspace{0.2cm}

When text uses BOTH a generic relation AND a name for the same person, LINK THEM!

\textbf{Example:}
\begin{itemize}
    \item Input: ``My roommate Emily got married. She works at Google.''
    \item Correct: ``Emily (the user's roommate) got married. She works at Google.''
    \item Wrong: Treating ``my roommate'' and ``Emily'' as separate entities
\end{itemize}
\end{tcolorbox}

\subsection{Opinion Formation Prompt (CARA)}
\label{sec:appendix-opinion-formation}

The opinion formation prompt is used during the reflect operation to extract and form new opinions from generated responses.

\begin{tcolorbox}[
    colback=teal!5!white,
    colframe=teal!60!black,
    title=Opinion Formation System Prompt,
    fonttitle=\bfseries,
    coltitle=white,
    colbacktitle=teal!60!black,
    boxrule=1.0pt,
    arc=1.5mm,
    enhanced,
    breakable
]
\small
\textbf{User Prompt:}

Extract any NEW opinions or perspectives from the answer below and rewrite them in FIRST-PERSON as if YOU are stating the opinion directly.

\vspace{0.2cm}

\textbf{ORIGINAL QUESTION:} \\
\texttt{\{query\}}

\vspace{0.2cm}

\textbf{ANSWER PROVIDED:} \\
\texttt{\{text\}}

\vspace{0.2cm}

Your task: Find opinions in the answer and rewrite them AS IF YOU ARE THE ONE SAYING THEM.

An opinion is a judgment, viewpoint, or conclusion that goes beyond just stating facts.

\vspace{0.2cm}

\textbf{IMPORTANT:} Do NOT extract statements like:
\begin{itemize}
    \item ``I don't have enough information''
    \item ``The facts don't contain information about X''
    \item ``I cannot answer because...''
\end{itemize}

ONLY extract actual opinions about substantive topics.

\vspace{0.2cm}

\rule{\linewidth}{0.4pt}

\textbf{CRITICAL FORMAT REQUIREMENTS:}

\rule{\linewidth}{0.4pt}

\vspace{0.2cm}

\textbf{1)} ALWAYS start with first-person phrases: ``I think...'', ``I believe...'', ``In my view...'', ``I've come to believe...'', ``Previously I thought... but now...''

\textbf{2)} NEVER use third-person: Do NOT say ``The speaker thinks...'' or ``They believe...'' - always use ``I''

\textbf{3)} Include the reasoning naturally within the statement

\textbf{4)} Provide a confidence score (0.0 to 1.0)

\vspace{0.2cm}

\textbf{CORRECT Examples (First-Person):}
\begin{itemize}
    \item ``I think Alice is more reliable because she consistently delivers on time and writes clean code''
    \item ``Previously I thought all engineers were equal, but now I feel that experience and track record really matter''
    \item ``I believe reliability is best measured by consistent output over time''
    \item ``I've come to believe that track records are more important than potential''
\end{itemize}
\end{tcolorbox}

\subsection{Observation Generation Prompt (TEMPR)}
\label{sec:appendix-observation-generation}

The observation generation prompt synthesizes factual observations about entities from multiple underlying facts without behavioral profile influence.

\begin{tcolorbox}[
    colback=teal!5!white,
    colframe=teal!60!black,
    title=Observation Generation System Prompt,
    fonttitle=\bfseries,
    coltitle=white,
    colbacktitle=teal!60!black,
    boxrule=1.0pt,
    arc=1.5mm,
    enhanced,
    breakable
]
\small
\textbf{System Message:}

You are an objective observer synthesizing facts about an entity. Generate clear, factual observations without opinions or behavioral profile influence. Be concise and accurate.

\vspace{0.2cm}

\rule{\linewidth}{0.4pt}

\textbf{User Prompt:}

Based on the following facts about ``\texttt{\{entity\_name\}}'', generate a list of key observations.

\vspace{0.2cm}

\textbf{FACTS ABOUT \texttt{\{ENTITY\_NAME\}}:} \\
\texttt{\{facts\_text\}}

\vspace{0.2cm}

Your task: Synthesize the facts into clear, objective observations about \texttt{\{entity\_name\}}.

\vspace{0.2cm}

\textbf{GUIDELINES:}
\begin{enumerate}
    \item Each observation should be a factual statement about \texttt{\{entity\_name\}}
    \item Combine related facts into single observations where appropriate
    \item Be objective - do not add opinions, judgments, or interpretations
    \item Focus on what we KNOW about \texttt{\{entity\_name\}}, not what we assume
    \item Include observations about: identity, characteristics, roles, relationships, activities
    \item Write in third person (e.g., ``John is...'' not ``I think John is...'')
    \item If there are conflicting facts, note the most recent or most supported one
\end{enumerate}

\vspace{0.2cm}

\textbf{EXAMPLES of good observations:}
\begin{itemize}
    \item ``John works at Google as a software engineer''
    \item ``John is detail-oriented and methodical in his approach''
    \item ``John collaborates frequently with Sarah on the AI project''
    \item ``John joined the company in 2023''
\end{itemize}

\textbf{EXAMPLES of bad observations (avoid these):}
\begin{itemize}
    \item ``John seems like a good person'' (opinion/judgment)
    \item ``John probably likes his job'' (assumption)
    \item ``I believe John is reliable'' (first-person opinion)
\end{itemize}

\vspace{0.2cm}

Generate 3-7 observations based on the available facts. If there are very few facts, generate fewer observations.
\end{tcolorbox}

\subsection{LongMemEval Judge Prompts}
\label{sec:appendix-judge-prompts}

The judge prompts are used in the LongMemEval benchmark to evaluate whether model responses are correct. Different prompts are used for different question types.

\subsubsection{Single-Session and Multi-Session Questions}

\begin{tcolorbox}[
    colback=teal!5!white,
    colframe=teal!60!black,
    title=Judge Prompt: Single/Multi-Session Questions,
    fonttitle=\bfseries,
    coltitle=white,
    colbacktitle=teal!60!black,
    boxrule=1.0pt,
    arc=1.5mm,
    enhanced,
    breakable
]
\small
I will give you a question, a correct answer, and a response from a model. Please answer yes if the response contains the correct answer. Otherwise, answer no. If the response is equivalent to the correct answer or contains all the intermediate steps to get the correct answer, you should also answer yes. If the response only contains a subset of the information required by the answer, answer no.

\vspace{0.2cm}

\textbf{Question:} \texttt{\{question\}}

\textbf{Correct Answer:} \texttt{\{answer\}}

\textbf{Model Response:} \texttt{\{response\}}

\vspace{0.2cm}

Is the model response correct?

You may provide reasoning, but you MUST end your response with your final answer in the format: \textbackslash boxed\{yes\} or \textbackslash boxed\{no\}
\end{tcolorbox}

\subsubsection{Temporal Reasoning Questions}

\begin{tcolorbox}[
    colback=teal!5!white,
    colframe=teal!60!black,
    title=Judge Prompt: Temporal Reasoning Questions,
    fonttitle=\bfseries,
    coltitle=white,
    colbacktitle=teal!60!black,
    boxrule=1.0pt,
    arc=1.5mm,
    enhanced,
    breakable
]
\small
I will give you a question, a correct answer, and a response from a model. Please answer yes if the response contains the correct answer. Otherwise, answer no. If the response is equivalent to the correct answer or contains all the intermediate steps to get the correct answer, you should also answer yes. If the response only contains a subset of the information required by the answer, answer no. In addition, do not penalize off-by-one errors for the number of days. If the question asks for the number of days/weeks/months, etc., and the model makes off-by-one errors (e.g., predicting 19 days when the answer is 18), the model's response is still correct.

\vspace{0.2cm}

\textbf{Question:} \texttt{\{question\}}

\textbf{Correct Answer:} \texttt{\{answer\}}

\textbf{Model Response:} \texttt{\{response\}}

\vspace{0.2cm}

Is the model response correct?

You may provide reasoning, but you MUST end your response with your final answer in the format: \textbackslash boxed\{yes\} or \textbackslash boxed\{no\}
\end{tcolorbox}

\subsubsection{Knowledge Update Questions}

\begin{tcolorbox}[
    colback=teal!5!white,
    colframe=teal!60!black,
    title=Judge Prompt: Knowledge Update Questions,
    fonttitle=\bfseries,
    coltitle=white,
    colbacktitle=teal!60!black,
    boxrule=1.0pt,
    arc=1.5mm,
    enhanced,
    breakable
]
\small
I will give you a question, a correct answer, and a response from a model. Please answer yes if the response contains the correct answer. Otherwise, answer no. If the response contains some previous information along with an updated answer, the response should be considered as correct as long as the updated answer is the required answer.

\vspace{0.2cm}

\textbf{Question:} \texttt{\{question\}}

\textbf{Correct Answer:} \texttt{\{answer\}}

\textbf{Model Response:} \texttt{\{response\}}

\vspace{0.2cm}

Is the model response correct?

You may provide reasoning, but you MUST end your response with your final answer in the format: \textbackslash boxed\{yes\} or \textbackslash boxed\{no\}
\end{tcolorbox}

\subsubsection{Preference Questions}

\begin{tcolorbox}[
    colback=teal!5!white,
    colframe=teal!60!black,
    title=Judge Prompt: Preference Questions,
    fonttitle=\bfseries,
    coltitle=white,
    colbacktitle=teal!60!black,
    boxrule=1.0pt,
    arc=1.5mm,
    enhanced,
    breakable
]
\small
I will give you a question, a rubric for desired personalized response, and a response from a model. Please answer yes if the response satisfies the desired response. Otherwise, answer no. The model does not need to reflect all the points in the rubric. The response is correct as long as it recalls and utilizes the user's personal information correctly.

\vspace{0.2cm}

\textbf{Question:} \texttt{\{question\}}

\textbf{Rubric:} \texttt{\{answer\}}

\textbf{Model Response:} \texttt{\{response\}}

\vspace{0.2cm}

Is the model response correct?

You may provide reasoning, but you MUST end your response with your final answer in the format: \textbackslash boxed\{yes\} or \textbackslash boxed\{no\}
\end{tcolorbox}

\subsubsection{Abstention Questions}

\begin{tcolorbox}[
    colback=teal!5!white,
    colframe=teal!60!black,
    title=Judge Prompt: Abstention Questions,
    fonttitle=\bfseries,
    coltitle=white,
    colbacktitle=teal!60!black,
    boxrule=1.0pt,
    arc=1.5mm,
    enhanced,
    breakable
]
\small
I will give you an unanswerable question, an explanation, and a response from a model. Please answer yes if the model correctly identifies the question as unanswerable. The model could say that the information is incomplete, or some other information is given but the asked information is not.

\vspace{0.2cm}

\textbf{Question:} \texttt{\{question\}}

\textbf{Explanation:} \texttt{\{answer\}}

\textbf{Model Response:} \texttt{\{response\}}

\vspace{0.2cm}

Does the model correctly identify the question as unanswerable?

You may provide reasoning, but you MUST end your response with your final answer in the format: \textbackslash boxed\{yes\} or \textbackslash boxed\{no\}
\end{tcolorbox}

\subsection{Structured Output Schemas}
\label{sec:appendix-schemas}

Hindsight uses Pydantic models to enforce structured output from LLM calls. This ensures reliable parsing and validation of extracted information.

\subsubsection{Fact Schema}

\begin{tcolorbox}[
    colback=teal!5!white,
    colframe=teal!60!black,
    title=Fact Extraction Schema (Pydantic),
    fonttitle=\bfseries,
    coltitle=white,
    colbacktitle=teal!60!black,
    boxrule=1.0pt,
    arc=1.5mm,
    enhanced,
    breakable
]
\small
\begin{verbatim}
class ExtractedFact(BaseModel):
    # Five required dimensions
    what: str  # Complete description with ALL specifics
    when: str  # Temporal info with day of week
    where: str # Specific locations, places, areas
    who: str   # All people/entities with relationships
    why: str   # Emotions, preferences, motivations

    # Classification
    fact_type: Literal["world", "experience", "opinion"]

    # Optional structured fields
    occurred_start: Optional[str] = None
    occurred_end: Optional[str] = None
    mentioned_at: Optional[str] = None
    entities: Optional[List[Entity]] = None
    causal_relations: Optional[List[CausalRelation]] = None

class Entity(BaseModel):
    text: str  # Named entity as it appears

class CausalRelation(BaseModel):
    target_fact_index: int  # Index of related fact
    relation_type: Literal[
        "causes", "caused_by", "enables", "prevents"
    ]
    strength: float  # 0.0 to 1.0
\end{verbatim}
\end{tcolorbox}

\subsubsection{Opinion Schema}

\begin{tcolorbox}[
    colback=teal!5!white,
    colframe=teal!60!black,
    title=Opinion Extraction Schema (Pydantic),
    fonttitle=\bfseries,
    coltitle=white,
    colbacktitle=teal!60!black,
    boxrule=1.0pt,
    arc=1.5mm,
    enhanced,
    breakable
]
\small
\begin{verbatim}
class Opinion(BaseModel):
    opinion: str  # First-person opinion statement
    confidence: float  # 0.0 to 1.0
    reasoning: str  # Why this opinion was formed

class OpinionExtractionResponse(BaseModel):
    opinions: List[Opinion] = Field(
        default_factory=list,
        description="List of opinions extracted from text"
    )
\end{verbatim}
\end{tcolorbox}

\subsubsection{Observation Schema}

\begin{tcolorbox}[
    colback=teal!5!white,
    colframe=teal!60!black,
    title=Observation Extraction Schema (Pydantic),
    fonttitle=\bfseries,
    coltitle=white,
    colbacktitle=teal!60!black,
    boxrule=1.0pt,
    arc=1.5mm,
    enhanced,
    breakable
]
\small
\begin{verbatim}
class Observation(BaseModel):
    observation: str  # Factual statement about entity

class ObservationExtractionResponse(BaseModel):
    observations: List[Observation] = Field(
        default_factory=list,
        description="List of observations about entity"
    )
\end{verbatim}
\end{tcolorbox}

\end{document}